\documentclass{article}

\usepackage[OT1]{fontenc}
\usepackage{iclr2027_conference,times}
\iclrfinalcopy

\usepackage{amsmath,amssymb,amsfonts,amsthm}
\usepackage{mathtools}
\usepackage{booktabs}
\usepackage{graphicx}
\usepackage{xcolor}
\usepackage{hyperref}
\usepackage{url}
\usepackage{enumitem}
\usepackage{algorithm}
\usepackage{algpseudocode}

\hypersetup{
  hidelinks,
  pdftitle={Model-Aware Schedules Improve Generation via Fiberwise Optimal Transport},
  pdfauthor={Luyi Jia, Boyan Zhang, Yilun Liu, Steffen Rulands}
}

\newcommand{\R}{\mathbb{R}}
\newcommand{\E}{\mathbb{E}}

\DeclareMathOperator{\Law}{Law}

\title{Model-Aware Schedules Improve Generation\\
via Fiberwise Optimal Transport}

\author{%
  Luyi Jia$^{1,*,\dagger}$ \quad
  Boyan Zhang$^{1,*}$ \quad
  Yilun Liu$^{2,3}$ \quad
  Steffen Rulands$^{1,\dagger}$\\[3pt]
  {\normalfont $^{1}$Arnold-Sommerfeld-Center for Theoretical Physics,}\\
  {\normalfont \hphantom{$^{1}$}Ludwig-Maximilians-Universit\"at M\"unchen, Munich, Germany}\\
  {\normalfont $^{2}$Institute of Informatics, Ludwig-Maximilians-Universit\"at M\"unchen, Munich, Germany}\\
  {\normalfont $^{3}$Munich Center for Machine Learning, Munich, Germany}\\[3pt]
  {\normalfont\small
    \texttt{luyi.jia@campus.lmu.de}\quad
    \texttt{boyan.zhang@campus.lmu.de}}\\
  {\normalfont\small
    \texttt{yilun.liu@tum.de}\quad
    \texttt{rulands@lmu.de}}\\[2pt]
  {\normalfont\small
    $^{*}$Equal contribution. \qquad
    $^{\dagger}$Corresponding authors.}
}

\begin{document}

\maketitle
\fancyhead{}
\renewcommand{\headrulewidth}{0pt}

\begin{abstract}
Diffusion and flow-matching schedules control the signal and noise
coefficients that mix data and noise along affine probability paths.
Minimizing a kinetic action defined on coefficient paths, motivated by optimal
transport, helps explain strong baselines but remains model-agnostic
and ignores prediction error.
Here we introduce a model-aware schedule construction based on
fiberwise optimal transport.
At a fixed time and state on the probability path, compatible signal/noise decompositions form an affine fiber.
We define a fiberwise prediction risk by averaging optimal-transport costs between the true and predictor-induced decompositions within these fibers.
On a fixed coefficient curve, combining this risk with
coefficient-path kinetic action yields a closed-form optimal time allocation.
This construction extends to general linear prediction targets, and
the risk profile can be estimated from an early baseline checkpoint.
We evaluate denoising diffusion probabilistic models (DDPMs) and flow
matching across prediction targets, training configurations,
risk-estimation checkpoints, datasets, and architectures. Our model-aware schedules consistently outperform strong baselines, including a \(38.6\%\) relative FID reduction for flow matching on CIFAR-10 at 16 function
evaluations. Each model-agnostic kinetic baseline determines its own
kinetic reference coordinate. In these coordinates, fiberwise-risk profiles from independently
trained models in different settings align closely after
normalization to unit area. The resulting schedule deformations used in training also align,
suggesting empirical universality across the evaluated models
and settings. Pretrained-checkpoint diagnostics extend this normalized-risk agreement
to larger conditional latent diffusion and 2-Rectified Flow (2-RF) models. A frozen analytic allocation template retains most of the
model-aware improvement without further risk estimation or
model-specific fitting.

\vspace{1ex}


\end{abstract}

\section{Introduction}

Diffusion and flow-matching models learn to generate samples from noise
\citep{ho2020denoising,lipman2023flow}.
During training, a data sample \(x_0\) and reference noise
\(\epsilon\) are commonly combined along an affine probability
path with state \(x_t=m_tx_0+s_t\epsilon\) at time $t$.
A schedule specifies how the signal and noise coefficients \(m_t,s_t\) evolve.
A kinetic action defined on the path of coefficients,
motivated by dynamical optimal transport
\citep{benamou2000computational},
integrates the squared speed of these coefficients over time.
On the respective fixed coefficient curves, its minimization yields
cosine-type diffusion schedules and the standard conditional
optimal-transport (Cond-OT) parameterization for flow matching
\citep{shaul2023kinetic,ikeda2025speedaccuracy}.
However, the coefficient-path kinetic action is model-agnostic and
ignores prediction error.
Its connection to optimal transport motivates a model-dependent
correction based on transport between signal/noise decompositions.

Here, we propose a model-aware schedule construction to improve
generation quality.
At fixed time \(t\) with state \(x_t=x\), the true and predictor-induced
signal/noise decompositions can differ even though both sum to
the same state.
All compatible decompositions form an affine fiber over \(x\).
We define a fiberwise prediction risk by averaging the
optimal-transport costs between the true and predictor-induced
decompositions within these fibers.
This transport is base-preserving: it measures decomposition
discrepancy at fixed \(x\), leaving the distribution \(p_t\) of \(x_t\)
unchanged.
We estimate the risk profile---the risk as a function of position
along the baseline coefficient curve---from a baseline checkpoint.
On a fixed coefficient curve, combining this risk with
coefficient-path kinetic action yields a closed-form optimal
time allocation.
The risk term favors less schedule time in higher-risk regions,
while the kinetic term penalizes excessively rapid traversal
(Figure~\ref{fig:fiberwise-risk-allocation}).

To compare risk profiles across models and settings, we use the
kinetic reference coordinates determined independently by the
model-agnostic kinetic baselines.
After unit-area normalization, fiberwise-risk profiles from
independently trained models align closely.
The resulting allocation deformations used in training also align.
This agreement is not assumed by the construction.
We use the shared risk shape and the closed-form allocation rule
to construct a frozen analytic allocation template
(Figure~\ref{fig:shared-risk-allocation};
Section~\ref{sec:shared-risk-allocation}).
These findings suggest that model-aware schedule design can
uncover reusable allocation structure across diffusion and
flow matching.

Our contributions are as follows:
\begin{itemize}[leftmargin=*]
    \item We formulate fiberwise prediction risk through base-preserving
    optimal transport on signal/noise decomposition fibers.
    Under the normalized symmetric product metric, the risk reduces to
    the \(s_t^2\)-weighted noise-prediction MSE and extends to general
    linear prediction targets, enabling direct risk estimation from the corresponding
    prediction errors.

    \item We introduce kinetic reference coordinates defined by
    model-agnostic kinetic baselines to compare risk profiles and
    allocation deformations across models and settings.
    We formulate model-aware scheduling as minimum coefficient-path
    kinetic action under a budget on schedule-time-integrated
    fiberwise prediction risk.
    The fixed-curve specialization yields a closed-form optimal
    time allocation for one-shot schedule construction, using a
    risk profile that can be estimated from an early baseline
    checkpoint.

    \item Across prediction targets, training configurations, datasets, and
    architectures, our schedules consistently improve strong DDPM and
    flow-matching baselines, including a \(38.6\%\) relative reduction in Fr\'echet Inception
    Distance (FID) \citep{heusel2017fid} at 16 function evaluations. Improvements also persist across multiple numerical solvers.

   \item Shared normalized risk shapes and allocation deformations
    in kinetic reference coordinates suggest empirical universality
    across the evaluated models and settings.
    Diagnostics on pretrained diffusion transformer (DiT) and 2-RF
    models extend the normalized-risk agreement to larger conditional
    latent models with distinct architectures and generative constructions.
    The frozen analytic allocation template retains most of the
    model-aware improvement.
\end{itemize}

\section{Related Work}
\label{sec:related-work}

\paragraph{Adjacent design choices at different levels.}

Diffusion and flow-matching models involve distinct design choices:
the inference-time solver grid; the training-time sampling distribution
and loss weighting; and the probability path and its traversal. Inference-time methods select
grids using discretization-error criteria, trajectory regularity, or
conditional-entropy change
\citep{sabour2024align,chen2024trajectory,stancevic2025entropic}.
BOSS uses dynamic programming for grid selection and also fine-tunes
the velocity field to its selected grid \citep{nguyen2024bellman}.
\citet{xu2026diagnosing} design training-time sampling distributions
and loss weights using estimated optimal loss values. InfoNoise adapts the training-time sampling distribution using an
online estimate of the conditional-entropy-rate profile
\citep{raya2026infonoise}. We change only the traversal of a fixed coefficient curve and use the
resulting schedule during retraining and sampling; the other
choices remain complementary.

\paragraph{Schedule and probability-path design.}

\citet{hang2025improved} design log-SNR importance sampling for continuous
diffusion training rather than the discrete DDPM setting evaluated here.
Constant Rate Scheduling reparameterizes variance-preserving (VP)
diffusion schedules to equalize a chosen rate of distributional change.
Its model-dependent procedure updates the schedule online, with training
experiments under \(\epsilon\)-prediction \citep{okada2026constant}.
Our construction instead estimates fiberwise prediction risk once from
a baseline checkpoint and holds the schedule fixed during retraining.
On prescribed paths, LayoutFlow tests a sine traversal for layout generation
\citep{guerreiro2024layoutflow}, while \citet{tsimpos2025optimal}
optimize traversal for a uniform-in-time spatial Lipschitz bound.
Concurrent work uses polynomial velocity profiles motivated by Euler's
local truncation error \citep{bondar2026velocity}.
Another concurrent approach uses loss-quantile traversal that slows
at high conditional-flow-matching loss \citep{tania2026difficulty}.
\citet{chen2025lipschitz} instead optimize interpolation coefficients
for averaged squared drift Lipschitzness.
Most of these methods address either diffusion or flow matching.
Our construction allocates time along a fixed coefficient curve
by balancing coefficient-path kinetics against
schedule-time-integrated fiberwise prediction risk.
The same formulation applies across general linear prediction
targets in both DDPM and flow matching.
Improvements also persist across multiple numerical solvers.
Empirical universality across the evaluated models and settings
supports a frozen analytic allocation template.
This template can be used directly as a schedule without
further risk estimation or model-specific fitting.

\paragraph{Kinetic and optimal-transport perspectives.}

Motivated by dynamical optimal transport
\citep{benamou2000computational}, \citet{shaul2023kinetic}
derive expressions for conditional and marginal kinetic energy
of affine probability paths.
With independent endpoints and matched second moments, the
normalized conditional energy is coefficient-path kinetic action,
minimized by Cond-OT.
They optimize the marginal kinetic energy over both the coefficient
curve and its traversal.
The resulting paths are evaluated in flow-matching experiments
using independent endpoints rather than minibatch OT coupling.
On ImageNet-64, the estimated optimum nearly coincides with the
standard Cond-OT coefficient path.
We instead minimize coefficient-path kinetic action over traversals of
a fixed coefficient curve under a budget on
schedule-time-integrated fiberwise prediction risk.
\citet{ikeda2025speedaccuracy} also recover the Cond-OT and exact cosine
parameterizations by minimizing coefficient-path kinetic action, with
the latter subject to the VP constraint.
Under their stated assumptions, they derive a speed--accuracy relation
between marginal Wasserstein action and a specific Wasserstein
generation-error sensitivity.
Base-preserving Wasserstein formulations restrict transport to
corresponding fibers over a shared base marginal
\citep{peszek2023heterogeneous,chemseddine2025conditional}.
We apply this structure to signal/noise decomposition fibers.
The resulting fiberwise prediction risk measures prediction
discrepancy at a fixed state and supplies the model-dependent term
in our schedule-design objective.

\section{Background} 
\label{sec:probability-paths-kinetics}

\subsection{Affine Probability Paths and Schedule Coordinates}
\label{sec:affine-paths-schedule-coordinates}

To describe diffusion and flow-matching schedules in a common framework,
we work with affine probability paths
\begin{equation}
\label{eq:affine-probability-path}
x_t=m_t x_0+s_t\epsilon,
\qquad
x_0\sim p_{\rm data},\quad
\epsilon\sim\mathcal N(0,I),
\qquad t\in[0,1],
\end{equation}
with marginals \(p_t=\Law(x_t)\).
A schedule specifies a parameterized coefficient path
\(t\mapsto(m_t,s_t)\). The underlying coefficient curve is the geometric
curve traced in the \((m,s)\)-plane, independent of traversal speed.
For VP/DDPM, write the squared signal coefficient as \(\bar\alpha_t\).
Then \(m_t=\sqrt{\bar\alpha_t}\) and
\(s_t=\sqrt{1-\bar\alpha_t}\), so \(m_t^2+s_t^2=1\).
Standard VP/DDPM uses independent \(x_0\) and \(\epsilon\).
Under our data-to-noise convention, the standard Cond-OT parameterization
is \(m_t=1-t\) and \(s_t=t\). 
With the endpoint marginals fixed, different couplings of data
and noise samples \((x_0,\epsilon)\) can change their joint law
and hence the prediction problem, while preserving the affine
coefficient representation above
\citep{ho2020denoising,lipman2023flow,pooladian2023multisample,tong2024improving}.

\subsection{Coefficient-Path Kinetics and Standard Baselines}
\label{sec:coefficient-path-kinetics}

Let \(\gamma(t)=(m_t,s_t)\) be a given coefficient path.
For a fixed endpoint pair \((x_0,\epsilon)\), differentiating
\(x_t=m_tx_0+s_t\epsilon\) with respect to \(t\) gives the path velocity
\(\dot x_t=\dot m_t x_0+\dot s_t\epsilon\).
Under standard regularity conditions, the expected pathwise kinetic action
\(\int_0^1 \E\|\dot x_t\|^2\,dt\) is an upper bound on the marginal
Wasserstein action of \(p_t\).
This marginal action integrates the squared \(2\)-Wasserstein speed
of \(p_t\) over time and upper-bounds the endpoint quadratic
transport cost.
In the settings covered by speed--accuracy relations, it also
upper-bounds a Wasserstein generation-error sensitivity
\citep{benamou2000computational,shaul2023kinetic,ikeda2025speedaccuracy}.
Prior kinetic analyses use the Euclidean action of the coefficient
path to explain strong baseline schedules
\citep{shaul2023kinetic,ikeda2025speedaccuracy}.
We adopt this action and refer to it as
\emph{coefficient-path kinetics}:
\begin{equation}
\label{eq:coefficient-path-kinetic-action}
    \mathcal J_{\rm kin}(\gamma)
    =
    \int_0^1
    \bigl(\dot m_t^2+\dot s_t^2\bigr)\,dt .
\end{equation}
For independent data/noise endpoints with matched second moments,
\(\mathcal J_{\rm kin}\) equals the normalized expected pathwise
kinetic action.
With unequal second moments or coupled endpoints, the pathwise
action is upper-bounded by a fixed multiple of
\(\mathcal J_{\rm kin}\)
(Appendix~\ref{app:pathwise-action-coefficient-kinetics}).

For fixed coefficient endpoints \((1,0)\) and \((0,1)\),
minimizing \(\mathcal J_{\rm kin}\) gives a model-agnostic kinetic
baseline with constant-speed traversal of the shortest admissible
coefficient curve.
We call the normalized time of this baseline the
\emph{kinetic reference coordinate} \(\tau\).
The resulting baselines are the standard Cond-OT parameterization
and, under the VP constraint, the exact cosine parameterization
\citep{shaul2023kinetic,ikeda2025speedaccuracy}.
For VP, the exact cosine parameterization is
\begin{equation}
\label{eq:exact-vp-parameterization}
    (m_\tau,s_\tau)
    =
    \left(
        \cos\frac{\pi\tau}{2},
        \sin\frac{\pi\tau}{2}
    \right).
\end{equation}
This parameterization has appeared in prior diffusion work
\citep{salimans2022progressive}.
Our DDPM baselines use the closely related offset-normalized cosine
schedule introduced by \citet{nichol2021improved},
which approximates this kinetic baseline
(Appendix~\ref{app:vp-kinetic-reference-coordinate}).

\section{Fiberwise Prediction Risk via Optimal Transport}
\label{sec:fiberwise-prediction-risk}

\begin{figure*}[t]
    \centering
    \includegraphics[
        width=0.98\textwidth
    ]{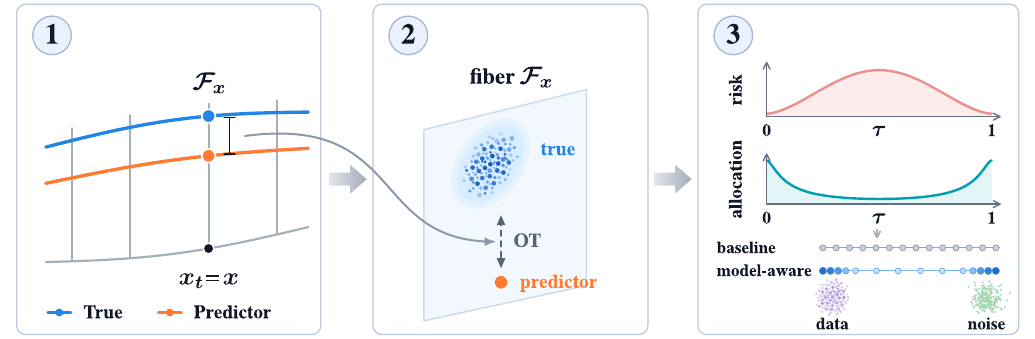}
    \caption{
        Fiberwise prediction risk and model-aware allocation.
        \textbf{1:} At \(x_t=x\), the true and predictor-induced decompositions
        differ vertically within \(\mathcal F_x\).
        \textbf{2:} The true conditional law and predictor-induced point mass
        define a base-preserving fiberwise transport cost, averaged over
        \(x\sim p_t\).
        \textbf{3:} The resulting risk and coefficient-path kinetics
        determine schedule-time allocation on the fixed curve,
        with profile shapes based on the empirical findings;
        marker spacing is schematic.
    }
    \label{fig:fiberwise-risk-allocation}
\end{figure*}

Section~\ref{sec:coefficient-path-kinetics} defines the kinetic
reference coordinate along each baseline coefficient curve. We now define the fiberwise prediction risk through base-preserving optimal
transport between the true and predictor-induced decompositions over each
fixed base state (Figure~\ref{fig:fiberwise-risk-allocation}, panels 1 and 2).

\subsection{Decomposition Fibers and Fiberwise Optimal Transport}
\label{sec:decomposition-fibers-risk}

To compare true and predictor-induced signal/noise decompositions
at a fixed state, we first describe the space of compatible
decompositions. Write \(x_t=u_t+n_t\), with \(u_t=m_tx_0\) and
\(n_t=s_t\epsilon\).
We define the affine fibers \(\mathcal F_x\) over states \(x\),
the decomposition bundle \(\mathcal B\) collecting these fibers,
and the projection \(\operatorname{pr}\) onto the base state:
\begin{equation}
\label{eq:decomposition-fibers-bundle-projection}
    \mathcal F_x=\{(u,n):u+n=x\},
    \qquad
    \mathcal B=\{(x,u,n):u+n=x\},
    \qquad
    \operatorname{pr}(x,u,n)=x .
\end{equation}
We equip the decomposition bundle with the normalized symmetric product
metric
\begin{equation}
\label{eq:normalized-product-metric}
    g_{\mathcal B}
    :=
    \frac12\bigl(\|du\|^2+\|dn\|^2\bigr).
\end{equation}
With the fiber coordinate \(z=(u-n)/2\), this becomes
\begin{equation}
\label{eq:base-fiber-metric-splitting}
    g_{\mathcal B}
    =
    \frac14\|dx\|^2+\|dz\|^2.
\end{equation}
This gives an orthogonal base--fiber splitting.
For a fixed endpoint pair, the true decomposition \((u_t,n_t)\)
evolves with time. The horizontal component of its velocity,
orthogonal to the fiber, has squared norm \(\|\dot x_t\|^2/4\).
In contrast, true and predictor-induced decompositions at the
same \((t,x)\) differ only in the vertical direction
(Appendix~\ref{app:orthogonal-base-fiber-splitting}).

For noise prediction, the true and predictor-induced decompositions are
\begin{equation}
\label{eq:true-predicted-decompositions}
    (u_t,n_t)
    =
    (m_tx_0,s_t\epsilon),
    \qquad
    (u_t^\theta,n_t^\theta)
    =
    \bigl(x_t-s_t\epsilon_\theta(x_t,t),
          s_t\epsilon_\theta(x_t,t)\bigr).
\end{equation}
At fixed \(t\), the observed state \(x_t=x\) may be compatible
with different true signal/noise decompositions.
Let \(\mu_t^x\) denote their conditional law given \(x_t=x\).
In contrast, a deterministic predictor returns a single
decomposition for the same \((x,t)\), inducing the point mass
\begin{equation}
\label{eq:predictor-induced-fiber-measure}
    \nu_t^{\theta,x}
    =
    \delta_{\zeta_\theta(x,t)},
    \qquad
    \zeta_\theta(x,t)
    =
    \bigl(x-s_t\epsilon_\theta(x,t),
          s_t\epsilon_\theta(x,t)\bigr)
    \in\mathcal F_x .
\end{equation}
With \(W_{2,\mathcal B}\) denoting the 2-Wasserstein distance on
\(\mathcal F_x\) induced by \(g_{\mathcal B}\), we define the
\emph{fiberwise prediction risk} by the base-preserving transport cost
\begin{equation}
\label{eq:fiberwise-prediction-risk}
    \mathcal W_{\rm fib}^2(t)
    :=
    \int
    W_{2,\mathcal B}^2
    \bigl(\mu_t^x,\nu_t^{\theta,x}\bigr)\,p_t(dx).
\end{equation}
Intuitively, \(\mathcal W_{\rm fib}^2(t)\) measures the average
squared distance between true and predicted signal/noise
decompositions that sum to the same observed state. Within each fiber, every true decomposition must be transported
to the single predicted decomposition.
The optimal-transport cost is therefore the conditional mean
squared vertical discrepancy.
Averaging over \(x\sim p_t\) gives the expected squared vertical
discrepancy under the joint law of \((x_0,\epsilon)\) inducing \(x_t\).
Writing \(z_t=(u_t-n_t)/2\) and
\(z_t^\theta=(u_t^\theta-n_t^\theta)/2\), we write this risk as
\begin{equation}
\label{eq:risk-noise-mse-identity}
    R(t)
    :=
    \mathcal W_{\rm fib}^2(t)
    =
    D_{\rm fib}^2(t)
    :=
    \E\|z_t-z_t^\theta\|^2
        =
    s_t^2e_t,
\end{equation}
with
\begin{equation}
    e_t :=\E\|\epsilon_\theta(x_t,t)-\epsilon\|^2 .
\end{equation}
This identity makes the risk directly estimable from prediction errors. The fiberwise transport defining \(R(t)\) compares decompositions
only within each fixed-\((t,x)\) fiber and therefore leaves
the base marginal \(p_t\) unchanged
(Appendix~\ref{app:fiberwise-ot-derivation}).

\subsection{General Linear Prediction Targets and Target-Coordinate Invariance}
\label{sec:general-linear-target-invariance}

To express the same fiberwise prediction risk for other
prediction targets, consider a general linear combination
of data and noise:
\begin{equation}
\label{eq:general-linear-target}
    y_t=a_tx_0+b_t\epsilon.
\end{equation}
A prediction \(y_\theta(x_t,t)\) induces the noise coordinate
\begin{equation}
\label{eq:induced-noise-prediction}
    \epsilon_\theta^{\rm ind}(x_t,t)
    =
    \frac{-a_tx_t+m_ty_\theta(x_t,t)}{\Delta_t},
\end{equation}
where $\Delta_t=m_tb_t-s_ta_t\neq0 $.
Since \(\epsilon_\theta^{\rm ind}-\epsilon =\frac{m_t}{\Delta_t}(y_\theta-y_t)\), the fiberwise risk can be written as
\begin{equation}
\label{eq:target-coordinate-risk}
    R(t)
    =
    s_t^2
    \E\|\epsilon_\theta^{\rm ind}(x_t,t)-\epsilon\|^2
    =
    \frac{m_t^2s_t^2}{\Delta_t^2}
    \E\|y_\theta(x_t,t)-y_t\|^2 .
\end{equation}
Thus changing the prediction target changes only the coordinate expression
of the same fiberwise risk.
This expression applies to the DDPM and flow-matching targets considered here.
This invariance does not imply matching risk profiles across separately
trained models, different settings, or coefficient curves.
The conditional and latent-state extension used for the pretrained-checkpoint
diagnostics is given in Appendix~\ref{app:pretrained-latent-diagnostics}.

For comparisons across models and settings, we express the
fiberwise risk as \(R(\tau)\) in the corresponding kinetic
reference coordinate.
Section~\ref{sec:model-aware-schedule-design} shows how this
profile enters schedule-time allocation.

\section{Model-Aware Schedule Design}
\label{sec:model-aware-schedule-design}

\subsection{Risk-Constrained Kinetic Formulation}
\label{sec:risk-constrained-kinetic-formulation}

To incorporate prediction risk into schedule design, we minimize
coefficient-path kinetic action subject to a budget on
schedule-time-integrated fiberwise prediction risk:
\begin{equation}
\label{eq:risk-constrained-schedule}
    \min_{\gamma}\;
    \mathcal J_{\rm kin}(\gamma)
    \qquad
    \text{subject to}
    \qquad
    \int_0^1 R(t)\,dt\le B .
\end{equation}
Under the conditions stated in
Section~\ref{sec:coefficient-path-kinetics} and
Appendix~\ref{app:pathwise-action-coefficient-kinetics},
this can also be viewed as minimizing a tractable upper bound
on the corresponding Wasserstein generation-error sensitivity
under a model-dependent risk budget. Introducing a Lagrange multiplier \(\lambda\ge0\) for the risk constraint
gives, up to an additive constant independent of the schedule,
the penalized objective
\begin{equation}
\label{eq:penalized-schedule-objective}
    \mathcal J_\lambda(\gamma)
    =
    \mathcal J_{\rm kin}(\gamma)
    +
    \lambda\int_0^1R(t)\,dt .
\end{equation}
For direct noise prediction, \(R(t)=s_t^2e_t\);
Section~\ref{sec:general-linear-target-invariance} gives the general-target
forms. The orthogonal base--fiber splitting in
Eq.~\eqref{eq:base-fiber-metric-splitting} separates changes in
the state from changes in its decomposition at a fixed state.
Under this splitting, a fixed multiple of
\(\mathcal J_{\rm kin}\) upper-bounds expected horizontal kinetic action
(Appendix~\ref{app:orthogonal-base-fiber-splitting}),
while \(R\) is a model-dependent vertical squared-distance potential.
These terms do not form a single dynamical optimal-transport action.

Rather than specifying \(B\), we parameterize the tradeoff by
\(\lambda\); the attained risk integral gives the corresponding budget.
We solve a one-shot fixed-curve problem that keeps the risk profile
estimated from a baseline checkpoint fixed and optimizes only traversal.
Empirically, re-estimating risk after model-aware training changes
the resulting time allocations only marginally, supporting this
one-shot construction (Appendix~\ref{app:shared-risk-allocation-agreement}).

\subsection{Fixed-Curve Model-Aware Allocation}
\label{sec:fixed-curve-model-aware-allocation}

To obtain an explicit allocation rule, we restrict
Eq.~\eqref{eq:risk-constrained-schedule} to traversals of a
fixed coefficient curve.
Let \(\gamma_0(\tau)\) denote the baseline coefficient path
in its kinetic reference coordinate.
With \(t=\Phi(\tau)\), the \emph{schedule-time allocation density} is
\begin{equation}
\label{eq:schedule-time-allocation-density}
    w(\tau)=\Phi'(\tau)>0,
    \qquad
    \int_0^1w(\tau)\,d\tau=1 .
\end{equation}
Thus \(w(\tau)\,d\tau\) is the schedule time assigned to a
reference interval of width \(d\tau\). Let \(L'(\tau)=\|\gamma_0'(\tau)\|\) denote the coefficient-space
speed of \(\gamma_0\), and write the previously estimated
fiberwise-risk profile as \(R(\tau)\).
We optimize \(w\) while holding this profile fixed.
The constrained problem becomes
\begin{equation}
\label{eq:fixed-curve-constrained-allocation}
    \begin{aligned}
    \min_{w>0}\quad&
        \int_0^1\frac{L'(\tau)^2}{w(\tau)}\,d\tau\\
    \text{subject to}\quad&
        \int_0^1R(\tau)w(\tau)\,d\tau\le B,
        \qquad
        \int_0^1w(\tau)\,d\tau=1 .
    \end{aligned}
\end{equation}
Introducing the corresponding multiplier \(\lambda\ge0\) gives the
penalized form
\begin{equation}
\label{eq:fixed-curve-penalized-allocation}
    \mathcal J_\lambda[w]
    =
    \int_0^1
    \left[
        \frac{L'(\tau)^2}{w(\tau)}
        +
        \lambda R(\tau)w(\tau)
    \right]d\tau .
\end{equation}
The Karush--Kuhn--Tucker (KKT) stationarity condition gives
the optimal allocation density in closed form:
\begin{equation}
\label{eq:optimal-allocation-density}
    w^\star(\tau)
    =
    \frac{L'(\tau)}
         {\sqrt{\eta+\lambda R(\tau)}} ,
\end{equation}
where \(\eta\) enforces normalization.
This rule converts an estimated risk profile into a time
allocation along the prescribed coefficient curve. The derivation, discrete allocation rule, and related invariance
properties are given in Appendix~\ref{app:fixed-curve-optimization}.

With \(L'(\tau)\) constant in the kinetic reference coordinate, higher values of
\(R(\tau)\) mean less schedule time when \(\lambda>0\).
The kinetic term penalizes arbitrarily rapid traversal.
Hence \(R\) and \(\lambda\) jointly determine the allocation
deformation: the change in schedule-time allocation relative to
the baseline traversal
(Figure~\ref{fig:fiberwise-risk-allocation}, panel 3).
Agreement in normalized risk-profile shapes alone does not
guarantee matching allocation deformations:
Eq.~\eqref{eq:optimal-allocation-density} also depends on the
absolute risk scale and on \(\lambda\) relative to the kinetic term.

\subsection{Instantiation for Diffusion and Flow Matching}
\label{sec:diffusion-flow-matching-instantiation}

For DDPM, we retain the standard offset-normalized cosine schedule as a
practical approximation to the VP kinetic baseline.
We deform only its traversal, using baseline time as the practical
VP reference coordinate.
Appendix~\ref{app:vp-kinetic-reference-coordinate} gives the coordinate
audit.

For flow matching, the standard Cond-OT time parameter is already the
exact reference coordinate along
\(\gamma_0(\tau)=(1-\tau,\tau)\). Under \(t=\Phi(\tau)\),
\begin{equation}
\label{eq:reparameterized-fm-path-target}
    x_t=(1-\tau)x_0+\tau\epsilon,
    \qquad
    \frac{dx_t}{dt}
    =
    \frac{\epsilon-x_0}{w(\tau)},
    \qquad
    \tau=\Phi^{-1}(t).
\end{equation}
Thus reparameterization changes both the state--time map and the linear
velocity prediction target, rather than merely the training-time sampling
distribution.
Appendix~\ref{app:model-aware-pipeline} gives implementation details
for both constructions.

\section{Experiments}
\label{sec:experiments}

\subsection{Experimental Setup}
\label{sec:experimental-setup}

We evaluate unconditional generation on CIFAR-10 and ImageNet-64
\citep{krizhevsky2009cifar,chrabaszcz2017downsampled} using U-Net
backbones and include a U-ViT-S/2 architecture-family control on CIFAR-10.
Primary DDPM experiments use \(1000\) timesteps and
\(\epsilon\)-prediction.
Our primary flow-matching setting is conditional flow matching with
minibatch OT endpoint coupling (OT-CFM), using a uniform
training-time sampling distribution and velocity prediction. Baseline and model-aware models are trained from scratch under
their respective schedules
(Appendix~\ref{app:model-aware-pipeline}).
In each FID comparison, all other training and sampling settings
are the same, and results are reported at matched training stages.

DDPM and flow matching use distinct U-Net implementations and
optimization recipes on CIFAR-10, but the same U-Net architecture
on ImageNet-64. The U-ViT control retains
the CIFAR-10 DDPM setup and constructs its schedule from its own baseline risk profile. Additional
controls vary the DDPM target, flow-matching endpoint coupling, and the training-time sampling distribution
(Appendices~\ref{app:model-aware-pipeline},
\ref{app:architectures-training-evaluation},
and~\ref{app:fm-coupling-time-sampling}).
Separately, we perform risk-profile and allocation diagnostics on public
DiT-XL/2 and pre-distillation InstaFlow 2-RF checkpoints, without retraining
or evaluating a modified generative schedule
(Appendix~\ref{app:pretrained-latent-diagnostics}).

Following established evaluation practice
\citep{shaul2023kinetic,sabour2024align,okada2026constant},
we compare generation quality using 50,000-sample FID at matched
numbers of function evaluations (NFE).
We define
\(\Delta\mathrm{FID}=\mathrm{FID}_{\mathrm{baseline}}
-\mathrm{FID}_{\mathrm{model\text{-}aware}}\), with positive values indicating
improvement. Main CIFAR-10 results give mean and standard deviation over three
pre-specified paired seeds; ImageNet-64 uses one paired seed. All other
single-seed CIFAR-10 studies use the same seed selected in advance from this set.

Coarse FID sweeps in the two primary CIFAR-10 settings set
\(\lambda=220\) for DDPM and \(\lambda=450\) for flow matching.
These sweeps precede the cross-system risk--allocation analysis.
We reuse the corresponding values without retuning across
prediction targets, endpoint couplings, training-time sampling
distributions, risk-estimation checkpoints, datasets, and
architectures
(Appendix~\ref{app:training-stage-tradeoff-sensitivity}).

\begin{table}[t]
    \caption{
        CIFAR-10 DDPM FID at epoch 400 with
        \(\epsilon\)-prediction and \(\lambda=220\). Values are mean \(\pm\) standard deviation over
        three paired training seeds.
    }
    \label{tab:ddpm-cifar-main}
    \centering
    \small
    \setlength{\tabcolsep}{6pt}
    \begin{tabular}{lccccc}
        \toprule
        & &
        \multicolumn{1}{c}{\textbf{Baseline}} &
        \multicolumn{3}{c}{\textbf{Model-aware (Ours)}} \\
        \cmidrule(lr){3-3}
        \cmidrule(lr){4-6}
    
        Sampler
        & NFE
        & FID~\(\downarrow\)
        & FID~\(\downarrow\)
        & \(\Delta\)FID~\(\uparrow\)
        & \%Impr.~\(\uparrow\) \\
        \midrule
        DPM++3M
        & 16
        & \(9.64 \pm 0.26\)
        & \(8.06 \pm 0.16\)
        & \(1.58 \pm 0.11\)
        & \(\mathbf{16.4 \pm 0.7\%}\) \\
        DPM++3M
        & 32
        & \(7.69 \pm 0.26\)
        & \(6.79 \pm 0.27\)
        & \(0.91 \pm 0.12\)
        & \(11.8 \pm 1.5\%\) \\
        DPM++3M
        & 64
        & \(6.83 \pm 0.24\)
        & \(6.19 \pm 0.27\)
        & \(0.64 \pm 0.07\)
        & \(9.3 \pm 1.2\%\) \\
        \midrule
        DDIM
        & 16
        & \(11.27 \pm 0.19\)
        & \(9.98 \pm 0.23\)
        & \(1.29 \pm 0.20\)
        & \(11.4 \pm 1.7\%\) \\
        DPM++2M
        & 16
        & \(9.81 \pm 0.26\)
        & \(8.29 \pm 0.14\)
        & \(1.51 \pm 0.14\)
        & \(15.4 \pm 1.1\%\) \\
        \bottomrule
    \end{tabular}
\end{table}

\begin{table}[t]
    \caption{
        Unconditional ImageNet-64 DDPM FID with
        \(\epsilon\)-prediction at the 800k checkpoint and
        \(\lambda=220\).
    }
    \label{tab:ddpm-imagenet-main}
    \centering
    \small
    \setlength{\tabcolsep}{12pt}
    \begin{tabular}{lccccc}
    
        \toprule
        & &
        \multicolumn{1}{c}{\textbf{Baseline}} &
        \multicolumn{3}{c}{\textbf{Model-aware (Ours)}} \\
        \cmidrule(lr){3-3}
        \cmidrule(lr){4-6}
    
        Sampler
        & NFE
        & FID~\(\downarrow\)
        & FID~\(\downarrow\)
        & \(\Delta\)FID~\(\uparrow\)
        & \%Impr.~\(\uparrow\) \\
        \midrule

        DPM++3M
        & 16
        & \(25.01\)
        & \(23.26\)
        & \(1.75\)
        & \(\mathbf{7.0\%}\) \\
        DPM++3M
        & 32
        & \(20.59\)
        & \(20.13\)
        & \(0.46\)
        & \(2.3\%\) \\
        DPM++3M
        & 64
        & \(20.05\)
        & \(19.50\)
        & \(0.55\)
        & \(2.7\%\) \\
        \midrule
        DDIM
        & 16
        & \(27.46\)
        & \(26.55\)
        & \(0.91\)
        & \(3.3\%\) \\
        DPM++2M
        & 16
        & \(24.90\)
        & \(23.39\)
        & \(1.51\)
        & \(6.1\%\) \\
        \bottomrule
    \end{tabular}
\end{table}

\subsection{Main Results}
\label{sec:main-results}

\paragraph{DDPM.}
\label{sec:ddpm-cifar-main}
\label{sec:ddpm-imagenet-main}
Our model-aware schedule reduces FID in all five reported sampling
configurations on CIFAR-10 and all five on ImageNet-64.
On CIFAR-10, the FID reduction reaches \(16.4\%\) with DPM++3M
at 16 NFE (Table~\ref{tab:ddpm-cifar-main}).
Across the complete \(3\times3\) sampler--NFE grid, FID is lower
for each of the three paired seeds
(Appendix Table~\ref{tab:ddpm-cifar-full-grid}).
Cosine also outperforms the linear-\(\beta\) schedule across all nine
configurations
(Appendix Table~\ref{tab:ddpm-linear-schedule}).
ImageNet-64 reductions at 16 NFE range from \(0.91\) to \(1.75\) FID,
with DPM++3M also improved at 32 and 64 NFE
(Table~\ref{tab:ddpm-imagenet-main}).

\paragraph{Flow matching.}
\label{sec:fm-cifar-main}
\label{sec:fm-imagenet-main}
All four CIFAR-10 sampling configurations improve for each of the three paired seeds,
with mean relative FID reductions of \(6.9\%\)--\(38.6\%\). All seven ImageNet-64 configurations improve at
600k updates, including every Euler budget; midpoint and Heun3 improve by
\(24.7\%\)--\(29.8\%\)
(Tables~\ref{tab:fm-cifar-main} and~\ref{tab:fm-imagenet-main}).

\begin{table}[t]
    \centering
    \small
    \setlength{\tabcolsep}{6pt}
    \caption{
        CIFAR-10 flow-matching FID at epoch 900 with
        \(\lambda=450\).
        Values are mean \(\pm\) standard deviation over three paired training
        seeds.
    }
    \label{tab:fm-cifar-main}
    \begin{tabular}{lccccc}
    
    \toprule
    & &
    \multicolumn{1}{c}{\textbf{Baseline}} &
    \multicolumn{3}{c}{\textbf{Model-aware (Ours)}} \\
    \cmidrule(lr){3-3}
    \cmidrule(lr){4-6}

    Integrator
    & NFE
    & FID~\(\downarrow\)
    & FID~\(\downarrow\)
    & \(\Delta\)FID~\(\uparrow\)
    & \%Impr.~\(\uparrow\) \\
    \midrule
    
        Midpoint
        & 16
        & \(7.62 \pm 0.20\)
        & \(4.68 \pm 0.13\)
        & \(2.95 \pm 0.09\)
        & \(\mathbf{38.6 \pm 0.6\%}\) \\
        Midpoint
        & 32
        & \(5.52 \pm 0.10\)
        & \(4.72 \pm 0.10\)
        & \(0.79 \pm 0.09\)
        & \(14.4 \pm 1.6\%\) \\
        Midpoint
        & 64
        & \(4.45 \pm 0.07\)
        & \(4.15 \pm 0.09\)
        & \(0.31 \pm 0.11\)
        & \(6.9 \pm 2.3\%\) \\
        \midrule
        Heun3
        & 15
        & \(7.19 \pm 0.17\)
        & \(4.88 \pm 0.27\)
        & \(2.32 \pm 0.12\)
        & \(32.2 \pm 2.3\%\) \\
        \bottomrule
    \end{tabular}
\end{table}

\begin{table}[t]
    \centering
    \small
    \setlength{\tabcolsep}{12pt}
    \caption{
        Unconditional ImageNet-64 flow-matching FID at the 600k
        checkpoint with \(\lambda=450\).
    }
    \label{tab:fm-imagenet-main}
    \begin{tabular}{lccccc}
    
    \toprule
    & &
    \multicolumn{1}{c}{\textbf{Baseline}} &
    \multicolumn{3}{c}{\textbf{Model-aware (Ours)}} \\
    \cmidrule(lr){3-3}
    \cmidrule(lr){4-6}

    Integrator
    & NFE
    & FID~\(\downarrow\)
    & FID~\(\downarrow\)
    & \(\Delta\)FID~\(\uparrow\)
    & \%Impr.~\(\uparrow\) \\
    \midrule

        Midpoint
        & 16
        & \(41.52\)
        & \(30.11\)
        & \(11.40\)
        & \(27.5\%\) \\
        Midpoint
        & 32
        & \(38.75\)
        & \(29.08\)
        & \(9.67\)
        & \(25.0\%\) \\
        Midpoint
        & 64
        & \(36.99\)
        & \(27.87\)
        & \(9.12\)
        & \(24.7\%\) \\
        \midrule
        Heun3
        & 15
        & \(41.38\)
        & \(29.04\)
        & \(12.34\)
        & \(\mathbf{29.8\%}\) \\
        \midrule
        Euler
        & 16
        & \(40.29\)
        & \(36.50\)
        & \(3.79\)
        & \(9.4\%\) \\
        Euler
        & 32
        & \(38.47\)
        & \(32.26\)
        & \(6.21\)
        & \(16.1\%\) \\
        Euler
        & 64
        & \(37.33\)
        & \(30.34\)
        & \(6.99\)
        & \(18.7\%\) \\
        \bottomrule
    \end{tabular}
\end{table}

\subsection{Robustness and Ablations}
\label{sec:robustness-ablations}

\paragraph{Target and architecture transfer.}
\label{sec:prediction-target-robustness}
Using \(\lambda=220\) throughout, the resulting schedules improve all six
DDPM \(v\)-prediction sampling configurations by \(7.4\%\)--\(14.7\%\) and all five
U-ViT-S/2 configurations without architecture-specific retuning, including
a \(12.2\%\) reduction at 16 NFE (Appendix Tables~\ref{tab:ddpm-v-prediction-results}
and~\ref{tab:ddpm-uvit-results}).

\paragraph{Coupling and training-time sampling.}
\label{sec:fm-coupling-time-sampling}
Across matched 400-epoch settings with different endpoint
couplings and training-time sampling distributions, our model-aware
schedules reduce FID by \(31.3\%\)--\(39.4\%\).
The FID rankings are preserved between baseline and model-aware runs:
OT coupling remains better than independent coupling, while logit-normal,
uniform, and U-shaped sampling remain ordered from lowest to highest FID
\citep{esser2024scaling,lee2024improving}. The U-shaped control replaces
the uniform training-time sampling distribution with the RF++ U-shaped distribution, held
fixed within each baseline--model-aware pair
(Appendix Table~\ref{tab:fm-coupling-time-sampling}).

\paragraph{Training-stage, tradeoff-weight, and risk-estimation robustness.}
Gains persist at the tested training stages and for the tested
nonzero \(\lambda\) values in both model families
(Appendix~\ref{app:training-stage-tradeoff-sensitivity}). Using epoch-100 rather than late-stage risk estimates changes final FID  by only
\(0.02\) for DDPM and \(0.06\) for flow matching while preserving the
shared normalized risk shape
(Appendix Tables~\ref{tab:ddpm-risk-checkpoint-sensitivity},
\ref{tab:fm-risk-checkpoint-sensitivity}, and
\ref{tab:shared-risk-allocation-agreement}).

\paragraph{Finite-step integration.}
\label{sec:finite-step-integration}
For flow matching, finite-NFE behavior is integrator-dependent.
On CIFAR-10, model-aware schedules improve FID across all tested
higher-order integrators and budgets despite low-NFE discretization
effects, but worsen FID with Euler. On ImageNet-64, they improve FID
with Euler at all three budgets, indicating setting-specific degradation
rather than systematic incompatibility with first-order integration
(Table~\ref{tab:fm-imagenet-main};
Appendix~\ref{app:fm-finite-step-diagnostic}).

\subsection{Shared Risk-Profile Shapes and Allocation Deformations}
\label{sec:shared-risk-allocation}

\begin{figure*}[t]
    \centering
    \includegraphics[
        height=0.295\textwidth,
        keepaspectratio
    ]{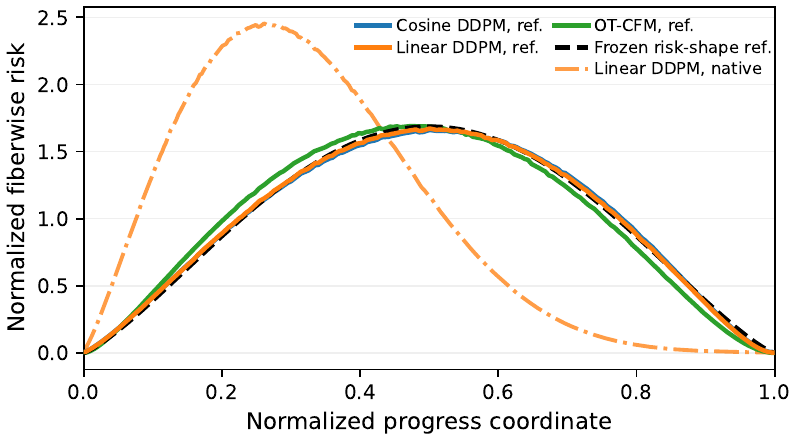}
    \hfill
    \includegraphics[
        height=0.295\textwidth,
        keepaspectratio
    ]{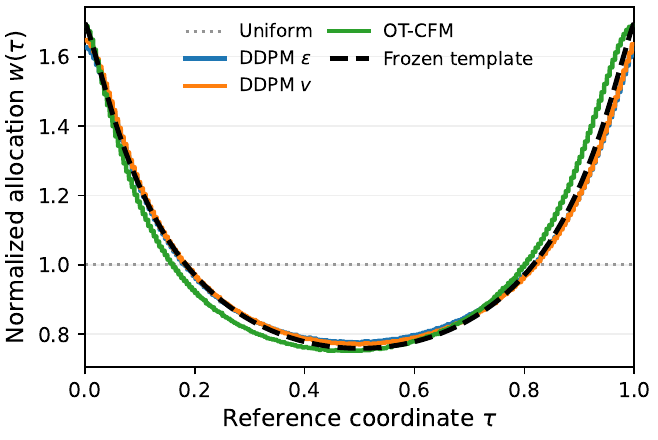}
    \caption{
        Shared risk-profile shapes and allocation deformations.
        \textbf{Left:} Fiberwise-risk profiles in kinetic reference coordinates,
        each normalized to unit area. The dash-dotted linear-DDPM profile instead
        uses native model time, and the dashed black curve is the frozen
        \(\sin^{5/4}(\pi\tau)\) risk-shape reference.
        \textbf{Right:} Model-aware allocation densities and the frozen analytic
        allocation template in kinetic reference coordinates.
    }
    \label{fig:shared-risk-allocation}
\end{figure*}

In the kinetic reference coordinates determined independently
by the model-agnostic baselines, the CIFAR-10 DDPM
\(\epsilon\)-prediction, DDPM \(v\)-prediction, and OT-CFM risk
profiles align closely after unit-area normalization.
Their allocation deformations also align closely, with more
schedule time near the endpoints than in the middle
(Figure~\ref{fig:shared-risk-allocation};
Appendix Table~\ref{tab:shared-risk-allocation-agreement}).
The DDPM and flow-matching \(\lambda\) values were selected
separately based on FID, not to align these deformations.

To characterize this agreement and examine its scope, we fit
analytic forms to the two primary CIFAR-10 baseline risk
profiles and their corresponding model-aware allocation
densities (DDPM \(\epsilon\)-prediction and OT-CFM).
The resulting \emph{frozen risk-shape reference} is the unit-area
normalization of \(\sin^{5/4}(\pi\tau)\).
The \emph{frozen analytic allocation template} uses the
reciprocal-square-root form of
Section~\ref{sec:fixed-curve-model-aware-allocation}:
\begin{equation}
\label{eq:main-analytic-allocation-template}
    w_{\rm ana}(\tau)
    \propto
    \left(1+4\sin^{5/4}(\pi\tau)\right)^{-1/2}.
\end{equation}
We apply these frozen reference profiles to the remaining
settings without refitting.
We compare unit-area risk profiles with the risk-shape reference
and allocation densities with the analytic template, using
Pearson correlation, relative \(L^2\) error, and total variation
(TV)
(Appendix~\ref{app:shared-risk-allocation-agreement}).

Appendix~\ref{app:shared-risk-allocation-agreement} extends both
findings across targets, checkpoints, architectures, datasets,
endpoint couplings, and flow-matching training-time sampling
distributions.
Using pretrained checkpoints without retraining, we also observe
the shared normalized risk shape in larger conditional latent
DiT-XL/2 and InstaFlow 2-RF models
(Appendices~\ref{app:pretrained-latent-diagnostics}
and~\ref{app:shared-risk-allocation-agreement}).

The linear-\(\beta\) DDPM control recovers this shared risk shape
in the VP kinetic reference coordinate, but not in native model time.
Linear and cosine DDPM follow the same VP coefficient curve but
assign different native model times to the same coefficient pairs.
At each native model time \(t\), the linear-DDPM risk \(R(t)\)
is evaluated at the corresponding coefficient pair \((m_t,s_t)\).
We express the profile in the practical VP reference coordinate
by matching each risk value to the cosine-baseline time with
the same coefficient pair.
In native model time, the linear-DDPM profile agrees weakly
with the risk-shape reference
(Pearson \(0.3616\), TV \(38.00\%\)).
In the practical VP reference coordinate, however, the linear-DDPM
profile closely matches the risk-shape reference
(Pearson \(0.9994\), TV \(0.84\%\);
Appendix~\ref{app:shared-risk-allocation-agreement}). In the corresponding kinetic reference coordinates, risk profiles
re-estimated from the primary CIFAR-10 DDPM and OT-CFM
model-aware checkpoints retain the shared normalized risk shape.
The recomputed allocations also retain the shared deformation
(Appendix~\ref{app:shared-risk-allocation-agreement}).

Overall, these results suggest empirical universality across the
evaluated models and settings.
The reference coordinate, normalized risk shape, and allocation
deformation connect the empirical comparison to our formulation.
Coefficient-path kinetics determines the reference coordinate,
and fiberwise optimal transport defines the model-dependent risk.
Their fixed-curve tradeoff determines the allocation deformation.

On CIFAR-10, the analytic template recovers
\(98.1\%\), \(98.0\%\), and \(75.8\%\) of the model-aware FID gains
for DDPM \(\epsilon\)-prediction, DDPM \(v\)-prediction, and OT-CFM,
respectively.
On ImageNet-64, it slightly outperforms the model-aware DDPM schedule
(\(109.7\%\) gain recovery) and recovers \(76.2\%\) of the
corresponding OT-CFM gain.
Each recovery fraction compares gains over the same baseline,
at the same training stage and with all other training and
sampling settings matched.
The analytic template thus distills rather than replaces the model-aware
construction: the latter identifies the shared normalized risk shape,
maps it to an allocation, and retains substantial additional
flow-matching gains
(Appendix~\ref{app:frozen-analytic-allocation-template-validation}).

An empirical Bayes decomposition for CIFAR-10 DDPM and
Independent-CFM attributes \(93.74\%\)--\(96.32\%\) of
\(\int_0^1 R(\tau)\,d\tau\) to predictor-dependent excess risk,
which also carries the shared normalized risk shape.
The small remaining Bayes-risk component is negatively
correlated with total risk
(Appendix~\ref{app:empirical-bayes-fiberwise-risk}).

\section{Discussion and Limitations}
\label{sec:discussion-limitations}

Our experiments show that combining coefficient-path kinetics
with fiberwise prediction risk to optimize traversal can improve
generation quality across models, training settings, and
sampling settings.
The shared risk shapes and allocation deformations suggest
empirical universality and motivate the analytic template,
which provides a directly usable schedule without further
risk estimation or model-specific fitting.
The additional flow-matching gains from the model-aware
schedules highlight the value of setting-specific risk
information.

The shared profile shape and quantitative cross-system agreement remain
theoretically unexplained. An explanation may require understanding how
the coefficient curve and training dynamics jointly shape fiberwise risk
and the resulting allocation deformations. Developing a unified path--fiber optimal-transport formulation
alongside an analysis of training dynamics may offer a route
to this explanation. Extending the one-shot
construction beyond fixed coefficient curves remains future work. We focus training and FID experiments on unconditional CIFAR-10
and ImageNet-64 for controlled comparisons across models,
training settings, and sampling settings. Pretrained-checkpoint diagnostics extend the analysis to
large conditional latent DiT and 2-RF models. End-to-end evaluation requires retraining these models under the
resulting schedules and remains future work.

\section*{Acknowledgments}

We thank Si-Yuan Chen for providing access to an NVIDIA GeForce
RTX 3090 GPU used in some of our experiments.
We gratefully acknowledge support from the hessian.AI Service Center
(funded by the Federal Ministry of Research, Technology and Space,
BMFTR, grant no. 16IS22091) and the hessian.AI Innovation Lab
(funded by the Hessian Ministry for Digital Strategy and Innovation,
grant no. S-DIW04/0013/003).
This work receives support from the Munich Center for Machine Learning
(MCML) and the Program of China Scholarship Council
(Grant No.202508080292). Steffen Rulands is a member of the Center for Nanoscience (CeNS).
The funding bodies had no role in the methodological or experimental
design, the analysis or interpretation of the results, or the writing
of the manuscript.


\section*{AI use statement}
ChatGPT was used to assist with text editing, coding, the generation and refinement of Figure 1 and assist with literature research. ChatGPT was not involved in research ideation, the development of theoretical results, research methodology, experimental design,  or selection and verification of citations. All AI-assisted outputs were reviewed, revised where necessary, and verified by the authors.

\bibliography{references}

\begin{thebibliography}{43}
\providecommand{\natexlab}[1]{#1}
\providecommand{\url}[1]{\texttt{#1}}
\expandafter\ifx\csname urlstyle\endcsname\relax
  \providecommand{\doi}[1]{doi: #1}\else
  \providecommand{\doi}{doi: \begingroup \urlstyle{rm}\Url}\fi

\bibitem[Bao et~al.(2023)Bao, Nie, Xue, Cao, Li, Su, and Zhu]{bao2023uvit}
Fan Bao, Shen Nie, Kaiwen Xue, Yue Cao, Chongxuan Li, Hang Su, and Jun Zhu.
\newblock All are worth words: A {ViT} backbone for diffusion models.
\newblock In \emph{Proceedings of the IEEE/CVF Conference on Computer Vision and Pattern Recognition}, pp.\  22669--22679, 2023.
\newblock \doi{10.1109/CVPR52729.2023.02171}.
\newblock URL \url{https://openaccess.thecvf.com/content/CVPR2023/html/Bao_All_Are_Worth_Words_A_ViT_Backbone_for_Diffusion_Models_CVPR_2023_paper.html}.

\bibitem[Benamou \& Brenier(2000)Benamou and Brenier]{benamou2000computational}
Jean-David Benamou and Yann Brenier.
\newblock A computational fluid mechanics solution to the {Monge--Kantorovich} mass transfer problem.
\newblock \emph{Numerische Mathematik}, 84\penalty0 (3):\penalty0 375--393, 2000.
\newblock \doi{10.1007/s002110050002}.

\bibitem[Bondar(2026)]{bondar2026velocity}
Vitalii Bondar.
\newblock Velocity scheduled flow matching.
\newblock \emph{arXiv preprint arXiv:2607.11442}, 2026.
\newblock URL \url{https://arxiv.org/abs/2607.11442}.

\bibitem[Chemseddine et~al.(2025)Chemseddine, Hagemann, Steidl, and Wald]{chemseddine2025conditional}
Jannis Chemseddine, Paul Hagemann, Gabriele Steidl, and Christian Wald.
\newblock Conditional wasserstein distances with applications in bayesian {OT} flow matching.
\newblock \emph{Journal of Machine Learning Research}, 26\penalty0 (141):\penalty0 1--47, 2025.
\newblock URL \url{https://www.jmlr.org/papers/v26/24-0586.html}.

\bibitem[Chen et~al.(2024)Chen, Zhou, Wang, Shen, and Lyu]{chen2024trajectory}
Defang Chen, Zhenyu Zhou, Can Wang, Chunhua Shen, and Siwei Lyu.
\newblock On the trajectory regularity of {ODE}-based diffusion sampling.
\newblock In \emph{Proceedings of the 41st International Conference on Machine Learning}, volume 235 of \emph{Proceedings of Machine Learning Research}, pp.\  7905--7934. PMLR, 2024.
\newblock URL \url{https://proceedings.mlr.press/v235/chen24bm.html}.

\bibitem[Chen et~al.(2025)Chen, Vanden-Eijnden, and Xu]{chen2025lipschitz}
Yifan Chen, Eric Vanden-Eijnden, and Jiawei Xu.
\newblock Lipschitz-guided design of interpolation schedules in generative models.
\newblock \emph{arXiv preprint arXiv:2509.01629}, 2025.
\newblock URL \url{https://arxiv.org/abs/2509.01629}.

\bibitem[Chrabaszcz et~al.(2017)Chrabaszcz, Loshchilov, and Hutter]{chrabaszcz2017downsampled}
Patryk Chrabaszcz, Ilya Loshchilov, and Frank Hutter.
\newblock A downsampled variant of {ImageNet} as an alternative to the {CIFAR} datasets.
\newblock \emph{arXiv preprint arXiv:1707.08819}, 2017.
\newblock URL \url{https://arxiv.org/abs/1707.08819}.

\bibitem[Deng et~al.(2009)Deng, Dong, Socher, Li, Li, and Li]{deng2009imagenet}
Jia Deng, Wei Dong, Richard Socher, Li-Jia Li, Kai Li, and Fei-Fei Li.
\newblock {ImageNet}: A large-scale hierarchical image database.
\newblock In \emph{2009 IEEE Conference on Computer Vision and Pattern Recognition}, pp.\  248--255, 2009.
\newblock \doi{10.1109/CVPR.2009.5206848}.
\newblock URL \url{https://doi.org/10.1109/CVPR.2009.5206848}.

\bibitem[Dhariwal \& Nichol(2021)Dhariwal and Nichol]{dhariwal2021adm}
Prafulla Dhariwal and Alexander Nichol.
\newblock Diffusion models beat {GANs} on image synthesis.
\newblock In \emph{Advances in Neural Information Processing Systems}, volume~34, pp.\  8780--8794, 2021.
\newblock URL \url{https://proceedings.neurips.cc/paper/2021/hash/49ad23d1ec9fa4bd8d77d02681df5cfa-Abstract.html}.

\bibitem[Esser et~al.(2024)Esser, Kulal, Blattmann, Entezari, M{\"u}ller, Saini, Levi, Lorenz, Sauer, Boesel, Podell, Dockhorn, English, and Rombach]{esser2024scaling}
Patrick Esser, Sumith Kulal, Andreas Blattmann, Rahim Entezari, Jonas M{\"u}ller, Harry Saini, Yam Levi, Dominik Lorenz, Axel Sauer, Frederic Boesel, Dustin Podell, Tim Dockhorn, Zion English, and Robin Rombach.
\newblock Scaling rectified flow transformers for high-resolution image synthesis.
\newblock In \emph{Proceedings of the 41st International Conference on Machine Learning}, volume 235 of \emph{Proceedings of Machine Learning Research}, pp.\  12606--12633. PMLR, 2024.
\newblock URL \url{https://proceedings.mlr.press/v235/esser24a.html}.

\bibitem[Guerreiro et~al.(2024)Guerreiro, Inoue, Masui, Otani, and Nakayama]{guerreiro2024layoutflow}
Julian Jorge~Andrade Guerreiro, Naoto Inoue, Kento Masui, Mayu Otani, and Hideki Nakayama.
\newblock {LayoutFlow}: Flow matching for layout generation.
\newblock In \emph{Computer Vision -- ECCV 2024}, volume 15094 of \emph{Lecture Notes in Computer Science}, pp.\  56--72. Springer, 2024.
\newblock \doi{10.1007/978-3-031-72764-1_4}.
\newblock URL \url{https://doi.org/10.1007/978-3-031-72764-1_4}.

\bibitem[Hang et~al.(2025)Hang, Gu, Bao, Wei, Chen, Geng, and Guo]{hang2025improved}
Tiankai Hang, Shuyang Gu, Jianmin Bao, Fangyun Wei, Dong Chen, Xin Geng, and Baining Guo.
\newblock Improved noise schedule for diffusion training.
\newblock In \emph{Proceedings of the IEEE/CVF International Conference on Computer Vision}, pp.\  4796--4806, 2025.
\newblock \doi{10.1109/ICCV51701.2025.00456}.
\newblock URL \url{https://openaccess.thecvf.com/content/ICCV2025/html/Hang_Improved_Noise_Schedule_for_Diffusion_Training_ICCV_2025_paper.html}.

\bibitem[Heusel et~al.(2017)Heusel, Ramsauer, Unterthiner, Nessler, and Hochreiter]{heusel2017fid}
Martin Heusel, Hubert Ramsauer, Thomas Unterthiner, Bernhard Nessler, and Sepp Hochreiter.
\newblock {GANs} trained by a two time-scale update rule converge to a local {Nash} equilibrium.
\newblock In \emph{Advances in Neural Information Processing Systems}, volume~30, pp.\  6626--6637. Curran Associates, Inc., 2017.
\newblock URL \url{https://proceedings.neurips.cc/paper/2017/hash/8a1d694707eb0fefe65871369074926d-Abstract.html}.

\bibitem[Ho \& Salimans(2022)Ho and Salimans]{ho2022classifierfree}
Jonathan Ho and Tim Salimans.
\newblock Classifier-free diffusion guidance.
\newblock \emph{arXiv preprint arXiv:2207.12598}, 2022.
\newblock URL \url{https://arxiv.org/abs/2207.12598}.

\bibitem[Ho et~al.(2020)Ho, Jain, and Abbeel]{ho2020denoising}
Jonathan Ho, Ajay Jain, and Pieter Abbeel.
\newblock Denoising diffusion probabilistic models.
\newblock In Hugo Larochelle, Marc'Aurelio Ranzato, Raia Hadsell, Maria-Florina Balcan, and Hsuan-Tien Lin (eds.), \emph{Advances in Neural Information Processing Systems}, volume~33, pp.\  6840--6851. Curran Associates, Inc., 2020.
\newblock URL \url{https://proceedings.neurips.cc/paper_files/paper/2020/file/4c5bcfec8584af0d967f1ab10179ca4b-Paper.pdf}.

\bibitem[Ikeda et~al.(2025)Ikeda, Uda, Okanohara, and Ito]{ikeda2025speedaccuracy}
Kotaro Ikeda, Tomoya Uda, Daisuke Okanohara, and Sosuke Ito.
\newblock Speed-accuracy relations for diffusion models: Wisdom from nonequilibrium thermodynamics and optimal transport.
\newblock \emph{Physical Review X}, 15\penalty0 (3):\penalty0 031031, July 2025.
\newblock \doi{10.1103/x5vj-8jq9}.
\newblock URL \url{https://doi.org/10.1103/x5vj-8jq9}.

\bibitem[Krizhevsky(2009)]{krizhevsky2009cifar}
Alex Krizhevsky.
\newblock Learning multiple layers of features from tiny images.
\newblock Technical report, University of Toronto, 2009.
\newblock URL \url{https://www.cs.toronto.edu/~kriz/learning-features-2009-TR.pdf}.

\bibitem[Lee et~al.(2024)Lee, Lin, and Fanti]{lee2024improving}
Sangyun Lee, Zinan Lin, and Giulia Fanti.
\newblock Improving the training of rectified flows.
\newblock In \emph{Advances in Neural Information Processing Systems}, volume~37, pp.\  63082--63109. Curran Associates, Inc., 2024.
\newblock \doi{10.52202/079017-2014}.
\newblock URL \url{https://proceedings.neurips.cc/paper_files/paper/2024/hash/7343a5c976f8399880b695267f1f9e9f-Abstract-Conference.html}.

\bibitem[Lin et~al.(2024)Lin, Liu, Li, and Yang]{lin2024flawed}
Shanchuan Lin, Bingchen Liu, Jiashi Li, and Xiao Yang.
\newblock Common diffusion noise schedules and sample steps are flawed.
\newblock In \emph{Proceedings of the IEEE/CVF Winter Conference on Applications of Computer Vision}, pp.\  5404--5411. IEEE, 2024.
\newblock \doi{10.1109/WACV57701.2024.00532}.
\newblock URL \url{https://openaccess.thecvf.com/content/WACV2024/html/Lin_Common_Diffusion_Noise_Schedules_and_Sample_Steps_Are_Flawed_WACV_2024_paper.html}.

\bibitem[Lin et~al.(2014)Lin, Maire, Belongie, Hays, Perona, Ramanan, Doll{\'a}r, and Zitnick]{lin2014coco}
Tsung-Yi Lin, Michael Maire, Serge Belongie, James Hays, Pietro Perona, Deva Ramanan, Piotr Doll{\'a}r, and C.~Lawrence Zitnick.
\newblock Microsoft {COCO}: Common objects in context.
\newblock In \emph{Computer Vision -- ECCV 2014}, volume 8693 of \emph{Lecture Notes in Computer Science}, pp.\  740--755. Springer, 2014.
\newblock \doi{10.1007/978-3-319-10602-1_48}.
\newblock URL \url{https://doi.org/10.1007/978-3-319-10602-1_48}.

\bibitem[Lipman et~al.(2023)Lipman, Chen, Ben-Hamu, Nickel, and Le]{lipman2023flow}
Yaron Lipman, Ricky T.~Q. Chen, Heli Ben-Hamu, Maximilian Nickel, and Matthew Le.
\newblock Flow matching for generative modeling.
\newblock In \emph{The Eleventh International Conference on Learning Representations}, 2023.
\newblock URL \url{https://openreview.net/forum?id=PqvMRDCJT9t}.

\bibitem[Liu et~al.(2023)Liu, Gong, and Liu]{liu2023flow}
Xingchao Liu, Chengyue Gong, and Qiang Liu.
\newblock Flow straight and fast: Learning to generate and transfer data with rectified flow.
\newblock In \emph{The Eleventh International Conference on Learning Representations}, 2023.
\newblock URL \url{https://openreview.net/forum?id=XVjTT1nw5z}.

\bibitem[Liu et~al.(2024)Liu, Zhang, Ma, Peng, and Liu]{liu2024instaflow}
Xingchao Liu, Xiwen Zhang, Jianzhu Ma, Jian Peng, and Qiang Liu.
\newblock {InstaFlow}: One step is enough for high-quality diffusion-based text-to-image generation.
\newblock In \emph{The Twelfth International Conference on Learning Representations}, 2024.
\newblock URL \url{https://openreview.net/forum?id=1k4yZbbDqX}.

\bibitem[Lu et~al.(2025)Lu, Zhou, Bao, Chen, Li, and Zhu]{lu2025dpmsolverpp}
Cheng Lu, Yuhao Zhou, Fan Bao, Jianfei Chen, Chongxuan Li, and Jun Zhu.
\newblock {DPM-Solver++}: Fast solver for guided sampling of diffusion probabilistic models.
\newblock \emph{Machine Intelligence Research}, 22\penalty0 (4):\penalty0 730--751, 2025.
\newblock \doi{10.1007/s11633-025-1562-4}.
\newblock URL \url{https://doi.org/10.1007/s11633-025-1562-4}.

\bibitem[Nguyen et~al.(2024)Nguyen, Nguyen, and Nguyen]{nguyen2024bellman}
Bao Nguyen, Binh Nguyen, and Viet~Anh Nguyen.
\newblock Bellman optimal stepsize straightening of flow-matching models.
\newblock In \emph{The Twelfth International Conference on Learning Representations}, 2024.
\newblock URL \url{https://openreview.net/forum?id=Iyve2ycvGZ}.

\bibitem[Nichol \& Dhariwal(2021)Nichol and Dhariwal]{nichol2021improved}
Alexander~Quinn Nichol and Prafulla Dhariwal.
\newblock Improved denoising diffusion probabilistic models.
\newblock In \emph{Proceedings of the 38th International Conference on Machine Learning}, volume 139 of \emph{Proceedings of Machine Learning Research}, pp.\  8162--8171. PMLR, 2021.
\newblock URL \url{https://proceedings.mlr.press/v139/nichol21a.html}.

\bibitem[Okada et~al.(2026)Okada, Doi, Yoshihashi, Kataoka, and Tanaka]{okada2026constant}
Shuntaro Okada, Kenji Doi, Ryota Yoshihashi, Hirokatsu Kataoka, and Tomohiro Tanaka.
\newblock Constant rate scheduling: A general framework for optimizing diffusion noise schedule via distributional change.
\newblock \emph{Transactions on Machine Learning Research}, 2026.
\newblock URL \url{https://openreview.net/forum?id=Pjq6kdvMBj}.

\bibitem[Peebles \& Xie(2023)Peebles and Xie]{peebles2023dit}
William Peebles and Saining Xie.
\newblock Scalable diffusion models with transformers.
\newblock In \emph{Proceedings of the IEEE/CVF International Conference on Computer Vision}, pp.\  4195--4205, 2023.
\newblock \doi{10.1109/ICCV51070.2023.00387}.
\newblock URL \url{https://openaccess.thecvf.com/content/ICCV2023/html/Peebles_Scalable_Diffusion_Models_with_Transformers_ICCV_2023_paper.html}.

\bibitem[Peszek \& Poyato(2023)Peszek and Poyato]{peszek2023heterogeneous}
Jan Peszek and David Poyato.
\newblock Heterogeneous gradient flows in the topology of fibered optimal transport.
\newblock \emph{Calculus of Variations and Partial Differential Equations}, 62\penalty0 (9):\penalty0 258, 2023.
\newblock \doi{10.1007/s00526-023-02601-8}.

\bibitem[Pooladian et~al.(2023)Pooladian, Ben-Hamu, Domingo-Enrich, Amos, Lipman, and Chen]{pooladian2023multisample}
Aram-Alexandre Pooladian, Heli Ben-Hamu, Carles Domingo-Enrich, Brandon Amos, Yaron Lipman, and Ricky T.~Q. Chen.
\newblock Multisample flow matching: Straightening flows with minibatch couplings.
\newblock In \emph{Proceedings of the 40th International Conference on Machine Learning}, volume 202 of \emph{Proceedings of Machine Learning Research}, pp.\  28100--28127. PMLR, 2023.
\newblock URL \url{https://proceedings.mlr.press/v202/pooladian23a.html}.

\bibitem[Raya et~al.(2026)Raya, Nguyen, Batzolis, Takida, Stancevic, Murata, Lai, Mitsufuji, and Ambrogioni]{raya2026infonoise}
Gabriel Raya, Bac Nguyen, Georgios Batzolis, Yuhta Takida, Dejan Stancevic, Naoki Murata, Chieh-Hsin Lai, Yuki Mitsufuji, and Luca Ambrogioni.
\newblock Noise scheduling as information-guided allocation in diffusion training.
\newblock \emph{arXiv preprint arXiv:2602.18647}, 2026.
\newblock URL \url{https://arxiv.org/abs/2602.18647}.

\bibitem[Rombach et~al.(2022)Rombach, Blattmann, Lorenz, Esser, and Ommer]{rombach2022latent}
Robin Rombach, Andreas Blattmann, Dominik Lorenz, Patrick Esser, and Bj{\"o}rn Ommer.
\newblock High-resolution image synthesis with latent diffusion models.
\newblock In \emph{Proceedings of the IEEE/CVF Conference on Computer Vision and Pattern Recognition}, pp.\  10684--10695, 2022.
\newblock \doi{10.1109/CVPR52688.2022.01042}.
\newblock URL \url{https://openaccess.thecvf.com/content/CVPR2022/html/Rombach_High-Resolution_Image_Synthesis_With_Latent_Diffusion_Models_CVPR_2022_paper.html}.

\bibitem[Ronneberger et~al.(2015)Ronneberger, Fischer, and Brox]{ronneberger2015unet}
Olaf Ronneberger, Philipp Fischer, and Thomas Brox.
\newblock {U-Net}: Convolutional networks for biomedical image segmentation.
\newblock In \emph{Medical Image Computing and Computer-Assisted Intervention -- MICCAI 2015}, volume 9351 of \emph{Lecture Notes in Computer Science}, pp.\  234--241. Springer, 2015.
\newblock \doi{10.1007/978-3-319-24574-4_28}.
\newblock URL \url{https://doi.org/10.1007/978-3-319-24574-4_28}.

\bibitem[Sabour et~al.(2024)Sabour, Fidler, and Kreis]{sabour2024align}
Amirmojtaba Sabour, Sanja Fidler, and Karsten Kreis.
\newblock Align your steps: Optimizing sampling schedules in diffusion models.
\newblock In \emph{Proceedings of the 41st International Conference on Machine Learning}, volume 235 of \emph{Proceedings of Machine Learning Research}, pp.\  42947--42975. PMLR, 2024.
\newblock URL \url{https://proceedings.mlr.press/v235/sabour24a.html}.

\bibitem[Salimans \& Ho(2022)Salimans and Ho]{salimans2022progressive}
Tim Salimans and Jonathan Ho.
\newblock Progressive distillation for fast sampling of diffusion models.
\newblock In \emph{The Tenth International Conference on Learning Representations}, 2022.
\newblock URL \url{https://openreview.net/forum?id=TIdIXIpzhoI}.

\bibitem[Shaul et~al.(2023)Shaul, Chen, Nickel, Le, and Lipman]{shaul2023kinetic}
Neta Shaul, Ricky T.~Q. Chen, Maximilian Nickel, Matthew Le, and Yaron Lipman.
\newblock On kinetic optimal probability paths for generative models.
\newblock In \emph{Proceedings of the 40th International Conference on Machine Learning}, volume 202 of \emph{Proceedings of Machine Learning Research}, pp.\  30883--30907. PMLR, 2023.
\newblock URL \url{https://proceedings.mlr.press/v202/shaul23a.html}.

\bibitem[Song et~al.(2021)Song, Meng, and Ermon]{song2021ddim}
Jiaming Song, Chenlin Meng, and Stefano Ermon.
\newblock Denoising diffusion implicit models.
\newblock In \emph{The Ninth International Conference on Learning Representations}, 2021.
\newblock URL \url{https://openreview.net/forum?id=St1giarCHLP}.

\bibitem[Stancevic et~al.(2025)Stancevic, Handke, and Ambrogioni]{stancevic2025entropic}
Dejan Stancevic, Florian Handke, and Luca Ambrogioni.
\newblock Entropic time schedulers for generative diffusion models.
\newblock In \emph{Advances in Neural Information Processing Systems}, volume~38, 2025.
\newblock \doi{10.52202/085713-1474}.
\newblock URL \url{https://proceedings.neurips.cc/paper_files/paper/2025/hash/3ed10616ecdd776be283c0a45cf9332d-Abstract-Conference.html}.

\bibitem[Tania \& Khan(2026)Tania and Khan]{tania2026difficulty}
Airin~Akter Tania and Md~Raihan Khan.
\newblock Difficulty-calibrated interpolation paths for conditional flow matching.
\newblock \emph{arXiv preprint arXiv:2608.21286}, 2026.
\newblock URL \url{https://arxiv.org/abs/2608.21286}.

\bibitem[Tong et~al.(2024)Tong, Fatras, Malkin, Huguet, Zhang, Rector-Brooks, Wolf, and Bengio]{tong2024improving}
Alexander Tong, Kilian Fatras, Nikolay Malkin, Guillaume Huguet, Yanlei Zhang, Jarrid Rector-Brooks, Guy Wolf, and Yoshua Bengio.
\newblock Improving and generalizing flow-based generative models with minibatch optimal transport.
\newblock \emph{Transactions on Machine Learning Research}, 2024.
\newblock URL \url{https://openreview.net/forum?id=CD9Snc73AW}.

\bibitem[Tsimpos et~al.(2025)Tsimpos, Zhi, Zech, and Marzouk]{tsimpos2025optimal}
Panos Tsimpos, Ren Zhi, Jakob Zech, and Youssef Marzouk.
\newblock Optimal scheduling of dynamic transport.
\newblock In \emph{Proceedings of Thirty Eighth Conference on Learning Theory}, volume 291 of \emph{Proceedings of Machine Learning Research}, pp.\  5441--5505. PMLR, 2025.
\newblock URL \url{https://proceedings.mlr.press/v291/tsimpos25a.html}.

\bibitem[von Platen et~al.(2022)von Platen, Patil, Lozhkov, Cuenca, Lambert, Rasul, Davaadorj, Nair, Paul, Liu, Berman, Xu, and Wolf]{vonplaten2022diffusers}
Patrick von Platen, Suraj Patil, Anton Lozhkov, Pedro Cuenca, Nathan Lambert, Kashif Rasul, Mishig Davaadorj, Dhruv Nair, Sayak Paul, Steven Liu, William Berman, Yiyi Xu, and Thomas Wolf.
\newblock Diffusers: State-of-the-art diffusion models.
\newblock GitHub repository, 2022.
\newblock URL \url{https://github.com/huggingface/diffusers}.

\bibitem[Xu et~al.(2026)Xu, Luo, Wang, He, and Liu]{xu2026diagnosing}
Yixian Xu, Shengjie Luo, Liwei Wang, Di~He, and Chang Liu.
\newblock Diagnosing and improving diffusion models by estimating the optimal loss value.
\newblock In \emph{The Fourteenth International Conference on Learning Representations}, 2026.
\newblock URL \url{https://openreview.net/forum?id=X7JfjLKKLQ}.

\end{thebibliography}
\bibliographystyle{iclr2027_conference}

\appendix
\numberwithin{equation}{section}

\section{Coefficient-Path Kinetics and Reference-Coordinate Details}
\label{app:coefficient-path-kinetics-reference-details}

\subsection{Relating Pathwise Action to Coefficient-Path Kinetics
under General Endpoint Joint Laws}
\label{app:pathwise-action-coefficient-kinetics}

This appendix relates expected pathwise kinetic action to coefficient-path
kinetics for fixed endpoint joint laws. We first establish a general
constant-factor upper bound, then identify cases of exact proportionality.

\paragraph{General endpoint joint laws.}
For a fixed joint law of \((x_0,\epsilon)\) with finite second moments,
\begin{equation}
\label{eq:app-affine-path-velocity}
    \dot x_t
    =
    \dot m_t x_0+\dot s_t\epsilon .
\end{equation}
Let
\begin{equation}
\label{eq:endpoint-second-moments}
    A=\E\|x_0\|^2,\qquad
    B=\E\|\epsilon\|^2,\qquad
    C=\E\langle x_0,\epsilon\rangle .
\end{equation}
Then
\begin{equation}
\label{eq:gram-pathwise-energy}
    \E\|\dot x_t\|^2
    =
    A\dot m_t^2
    +2C\dot m_t\dot s_t
    +B\dot s_t^2
    =
    \dot\gamma(t)^\top
    G
    \dot\gamma(t),
\end{equation}
where
\begin{equation}
\label{eq:endpoint-gram-matrix}
    G
    =
    \begin{pmatrix}
        A & C\\
        C & B
    \end{pmatrix}.
\end{equation}
The matrix \(G\) is positive semidefinite. Let \(c_G>0\) denote its
largest eigenvalue. Integrating the quadratic-form bound gives
\begin{equation}
\label{eq:pathwise-action-upper-bound}
    \int_0^1 \E\|\dot x_t\|^2\,dt
    =
    \int_0^1 \dot\gamma(t)^\top G\dot\gamma(t)\,dt
    \le c_G\mathcal J_{\rm kin}(\gamma).
\end{equation}
For a fixed endpoint joint law, \(c_G\) is independent of traversal.
Minimizing \(\mathcal J_{\rm kin}\) therefore minimizes this upper bound,
but not necessarily the pathwise action itself.

\paragraph{Independent endpoints.}
For independent endpoints, as in standard VP/DDPM, the centered noise
gives \(C=0\), and the bound uses \(c_G=\max(A,B)\).
If their second moments also match (\(A=B\)), then
\begin{equation}
\label{eq:matched-moment-pathwise-action}
    \int_0^1 \E\|\dot x_t\|^2\,dt
    =
    A\mathcal J_{\rm kin}(\gamma).
\end{equation}
This recovers the normalized relation used in the main text.
Under the same conditions, \(\E\|x_t\|^2=A(m_t^2+s_t^2)\), so the VP
constraint preserves the common second-moment scale.

\paragraph{Fixed Cond-OT coefficient curve.}
On the fixed Cond-OT coefficient curve, write
\begin{equation}
\label{eq:cond-ot-traversal}
    m_t=1-r(t),
    \qquad
    s_t=r(t),
    \qquad
    r(0)=0,\quad r(1)=1.
\end{equation}
Then
\begin{equation}
\label{eq:cond-ot-pathwise-energy}
    \E\|\dot x_t\|^2
    =
    (A+B-2C)\dot r(t)^2,
\end{equation}
where
\begin{equation}
\label{eq:endpoint-displacement-second-moment}
    A+B-2C
    =
    \E\|x_0-\epsilon\|^2
    \ge 0 .
\end{equation}
Since \(\mathcal J_{\rm kin}(\gamma)=2\int_0^1\dot r(t)^2\,dt\),
the pathwise action satisfies
\begin{equation}
\label{eq:cond-ot-action-proportionality}
    \int_0^1 \E\|\dot x_t\|^2\,dt
    =
    (A+B-2C)
    \int_0^1 \dot r(t)^2\,dt
    =
    \frac{A+B-2C}{2}\mathcal J_{\rm kin}(\gamma).
\end{equation}
This identity holds for any fixed endpoint joint law. Along this curve,
endpoint moments and coupling change only the proportionality factor
relating pathwise action to coefficient-path kinetics.
For a fixed nondegenerate endpoint joint law with \(A+B-2C>0\),
replacing coefficient-path kinetics by pathwise action therefore only
rescales the relative weighting of action and risk.

\paragraph{Cond-OT optimality without fixing the coefficient curve.}
For any absolutely continuous coefficient path with
\(\gamma(0)=(1,0)\) and \(\gamma(1)=(0,1)\), the Cauchy--Schwarz
inequality gives
\begin{equation}
\label{eq:coefficient-action-global-bound}
    \mathcal J_{\rm kin}(\gamma)
    \ge
    \left\|\int_0^1\dot\gamma(t)\,dt\right\|^2
    =2.
\end{equation}
Applying the same inequality to each sample path and averaging gives
\begin{equation}
\label{eq:pathwise-action-global-bound}
    \int_0^1 \E\|\dot x_t\|^2\,dt
    \ge
    \E\left\|\int_0^1\dot x_t\,dt\right\|^2
    =
    \E\|\epsilon-x_0\|^2
    =A+B-2C.
\end{equation}
The standard Cond-OT coefficient path \(\gamma(t)=(1-t,t)\)
attains both bounds. Thus, for each fixed endpoint joint law, it
minimizes both coefficient-path kinetics and pathwise action when
only the coefficient endpoints are prescribed.

\subsection{Practical Approximation to the VP Kinetic Reference Coordinate}
\label{app:vp-kinetic-reference-coordinate}

For the VP quarter-circle coefficient curve, the exact VP kinetic
reference coordinate is
\begin{equation}
\label{eq:exact-vp-reference-coordinate}
    \tau_{\rm VP}(t)
    =
    \frac{2}{\pi}
    \arccos\sqrt{\bar\alpha(t)},
\end{equation}
with \(\tau_{\rm VP}=0\) at the data endpoint and
\(\tau_{\rm VP}=1\) at the noise endpoint.

The standard offset-normalized cosine schedule used for our DDPM
baselines parameterizes the same VP coefficient curve through
\begin{equation}
\label{eq:offset-normalized-cosine-schedule}
    \varphi(t)
    =
    \frac{\pi}{2}
    \frac{t+s_{\rm off}}{1+s_{\rm off}},
    \qquad
    \bar\alpha(t)
    =
    \frac{\cos^2\varphi(t)}
         {\cos^2\varphi(0)},
\end{equation}
where \(s_{\rm off}\) is the cosine offset.
The cosine phase \(\varphi(t)\) is affine in baseline time \(t\).
However, normalizing \(\cos^2\varphi(t)\) by \(\cos^2\varphi(0)\)
makes \(\tau_{\rm VP}(t)\) differ from \(t\).
We retain \(t\) as the practical VP reference coordinate.
For \(s_{\rm off}=0.008\),
\begin{equation}
\label{eq:vp-reference-coordinate-discrepancy}
    \left\|
        t-\tau_{\rm VP}(t)
    \right\|_\infty
    =
    6.996\times10^{-3}.
\end{equation}

Appendix~\ref{app:shared-risk-allocation-agreement} reports the
corresponding DDPM coordinate controls.
Using the exact rather than practical VP reference coordinate
changes the DDPM risk-shape comparisons only marginally.

\section{Decomposition-Bundle Geometry and Fiberwise Optimal-Transport Details}
\label{app:decomposition-bundle-geometry}

\subsection{Orthogonal Base--Fiber Splitting}
\label{app:orthogonal-base-fiber-splitting}

We give the tangent-space calculation behind
Section~\ref{sec:decomposition-fibers-risk}. On the decomposition bundle,
use the normalized symmetric product metric introduced in the main text,
\begin{equation}
\label{eq:app-normalized-product-metric}
    \|(\delta u,\delta n)\|_{\mathcal B}^2
    :=
    \frac12
    \left(
        \|\delta u\|^2+\|\delta n\|^2
    \right).
\end{equation}
With
\begin{equation}
\label{eq:base-fiber-coordinate-transform}
    z=\frac{u-n}{2},
    \qquad
    u=\frac{x}{2}+z,
    \qquad
    n=\frac{x}{2}-z,
\end{equation}
we have
\begin{equation}
\label{eq:base-fiber-tangent-transform}
    \delta u=\frac12\delta x+\delta z,
    \qquad
    \delta n=\frac12\delta x-\delta z,
\end{equation}
and hence
\begin{equation}
\label{eq:app-base-fiber-metric-splitting}
    \frac12
    \left(
        \|\delta u\|^2+\|\delta n\|^2
    \right)
    =
    \frac14\|\delta x\|^2+\|\delta z\|^2 .
\end{equation}
Equivalently, the vertical tangent space and its metric-orthogonal
complement are
\begin{equation}
\label{eq:vertical-horizontal-tangent-spaces}
    V=\ker d\operatorname{pr}=\{(a,-a)\},
    \qquad
    H=V^\perp=\{(b,b)\}.
\end{equation}
In the \((x,z)\) coordinates, these are
\begin{equation}
\label{eq:tangent-spaces-in-base-fiber-coordinates}
    V=\{(\delta x,\delta z):\delta x=0\},
    \qquad
    H=\{(\delta x,\delta z):\delta z=0\},
\end{equation}
so the tangent space splits orthogonally into base and fiber directions.

A base displacement \(\delta x\) has horizontal lift
\begin{equation}
\label{eq:horizontal-base-lift}
    \left(
        \frac{\delta x}{2},
        \frac{\delta x}{2}
    \right),
\end{equation}
with squared norm \(\frac14\|\delta x\|^2\). For the true lifted path
\(\ell_t=(x_t,u_t,n_t)\), the horizontal projection therefore satisfies
\begin{equation}
\label{eq:horizontal-energy-identity}
    \|\dot\ell_t^H\|_{\mathcal B}^2
    =
    \frac14\|\dot x_t\|^2 .
\end{equation}
Expected horizontal kinetic energy is therefore one quarter of
expected pathwise kinetic energy. This does not imply that the full lifted tangent
\(\dot\ell_t\) is horizontal: the true decomposition may also move in the
fiber direction as \(t\) varies. Rather, the splitting isolates the
base-motion component from
same-base decomposition discrepancies, which are purely vertical.
For the true and predictor-induced lifts, which share the same base state,
\begin{equation}
\label{eq:vertical-prediction-discrepancy}
    \E\|z_t-z_t^\theta\|^2
    =
    D_{\rm fib}^2(t),
\end{equation}
so the fiberwise prediction discrepancy is purely vertical.
Appendix~\ref{app:pathwise-action-coefficient-kinetics}
bounds the pathwise action by a fixed multiple of
\(\mathcal J_{\rm kin}\).
Expected horizontal kinetic action is one quarter of the pathwise
action and therefore satisfies the corresponding bound.
Our objective combines \(\mathcal J_{\rm kin}\) with a vertical
squared-distance potential; it is not the full kinetic action
of the lifted path.

\subsection{Full Fiberwise Optimal-Transport Derivation}
\label{app:fiberwise-ot-derivation}

This subsection gives the full disintegration argument behind the
base-preserving fiberwise optimal-transport construction in
Section~\ref{sec:decomposition-fibers-risk}. Recall the decomposition bundle
\begin{equation}
\label{eq:app-decomposition-bundle-projection}
    \mathcal B
    =
    \{(x,u,n)\in\R^d\times\R^d\times\R^d:u+n=x\},
    \qquad
    \operatorname{pr}(x,u,n)=x,
\end{equation}
with fiber
\begin{equation}
\label{eq:app-decomposition-fiber}
    \mathcal F_x
    =
    \operatorname{pr}^{-1}(x)
    =
    \{(u,n):u+n=x\}.
\end{equation}
At a fixed time \(t\), the true and predictor-induced lifts
\begin{equation}
\label{eq:true-predicted-lifts}
    \ell_t=(x_t,u_t,n_t),
    \qquad
    \ell_t^\theta=(x_t,u_t^\theta,n_t^\theta)
\end{equation}
are random points in \(\mathcal B\) satisfying
\begin{equation}
\label{eq:shared-base-projection}
    \operatorname{pr}(\ell_t)=\operatorname{pr}(\ell_t^\theta)=x_t.
\end{equation}
Their laws are
\begin{equation}
\label{eq:true-predicted-lift-laws}
    \mu_t
    =
    \Law(x_t,u_t,n_t),
    \qquad
    \nu_t^\theta
    =
    \Law(x_t,u_t^\theta,n_t^\theta),
\end{equation}
where
\begin{equation}
\label{eq:true-signal-noise-components}
    u_t=m_t x_0,\qquad n_t=s_t\epsilon,
\end{equation}
and
\begin{equation}
\label{eq:predicted-signal-noise-components}
    n_t^\theta=s_t\epsilon_\theta(x_t,t),
    \qquad
    u_t^\theta=x_t-n_t^\theta.
\end{equation}
Both measures project to the same base marginal
\(p_t=\Law(x_t)\), and therefore admit disintegrations over the same base
variable:
\begin{equation}
\label{eq:true-law-disintegration}
    \mu_t(dx,du,dn)
    =
    p_t(dx)\,\mu_t^x(du,dn),
\end{equation}
\begin{equation}
\label{eq:predicted-law-disintegration}
    \nu_t^\theta(dx,du,dn)
    =
    p_t(dx)\,\nu_t^{\theta,x}(du,dn).
\end{equation}

For each fixed base point \(x\), \(\mu_t^x\) is the conditional
decomposition law supported on \(\mathcal F_x\):
\begin{equation}
\label{eq:conditional-decomposition-law}
    \mu_t^x
    =
    \Law\bigl((m_t x_0,s_t\epsilon)\mid x_t=x\bigr).
\end{equation}
For a deterministic predictor,
\begin{equation}
\label{eq:conditional-predictor-dirac-law}
    \nu_t^{\theta,x}
    =
    \delta_{\zeta_\theta(x,t)},
\end{equation}
where
\begin{equation}
\label{eq:predictor-fiber-point}
    \zeta_\theta(x,t)
    =
    \bigl(
        x-s_t\epsilon_\theta(x,t),
        s_t\epsilon_\theta(x,t)
    \bigr)
    \in\mathcal F_x .
\end{equation}

Let
\begin{equation}
\label{eq:fiber-squared-distance}
    d_{\mathcal B}^2
    \bigl((u,n),(u',n')\bigr)
    =
    \frac12
    \bigl(\|u-u'\|^2+\|n-n'\|^2\bigr)
\end{equation}
denote the squared distance on each fiber induced by the normalized
symmetric product metric \(g_{\mathcal B}\).
We define the fiberwise prediction risk by
\begin{equation}
\label{eq:app-fiberwise-risk-definition}
    \mathcal W_{\rm fib}^2(t)
    :=
    \int
    W_{2,\mathcal B}^2
    \bigl(\mu_t^x,\nu_t^{\theta,x}\bigr)
    \,p_t(dx).
\end{equation}

Because \(\nu_t^{\theta,x}\) is a Dirac mass, its coupling with
\(\mu_t^x\) is unique and therefore optimal. Hence
\begin{equation}
\label{eq:dirac-fiber-transport-integral}
    W_{2,\mathcal B}^2
    \bigl(\mu_t^x,\nu_t^{\theta,x}\bigr)
    =
    \int_{\mathcal F_x}
    d_{\mathcal B}^2
    \bigl((u,n),\zeta_\theta(x,t)\bigr)
    \,\mu_t^x(du,dn).
\end{equation}
Equivalently,
\begin{equation}
\label{eq:dirac-fiber-transport-expectation}
    W_{2,\mathcal B}^2
    \bigl(\mu_t^x,\nu_t^{\theta,x}\bigr)
    =
    \E\left[
        d_{\mathcal B}^2
        \bigl(
            (u_t,n_t),
            (u_t^\theta,n_t^\theta)
        \bigr)
        \,\middle|\, x_t=x
    \right].
\end{equation}

Averaging over \(x\sim p_t\) gives
\begin{equation}
\label{eq:averaged-fiber-transport-cost}
    \mathcal W_{\rm fib}^2(t)
    =
    \frac{1}{2}
    \E\left[
        \|u_t-u_t^\theta\|^2
        +
        \|n_t-n_t^\theta\|^2
    \right].
\end{equation}
Since the true and predictor-induced decompositions lie in the same fiber,
\begin{equation}
\label{eq:shared-state-decomposition}
    u_t+n_t
    =
    u_t^\theta+n_t^\theta
    =
    x_t,
\end{equation}
and therefore
\begin{equation}
\label{eq:opposite-component-discrepancies}
    u_t-u_t^\theta
    =
    -(n_t-n_t^\theta).
\end{equation}
It follows that
\begin{equation}
\label{eq:app-fiberwise-risk-identity}
    \begin{aligned}
    R(t)
    &:=
    \mathcal W_{\rm fib}^2(t)
    =
    D_{\rm fib}^2(t)
    =
    \E\|n_t-n_t^\theta\|^2\\
    &=
    s_t^2
    \E\|\epsilon-\epsilon_\theta(x_t,t)\|^2
    =
    s_t^2 e_t .
    \end{aligned}
\end{equation}

The same quantity can equivalently be formulated as an optimal-transport
problem on the total lifted space. Write
\begin{equation}
\label{eq:lifted-transport-points}
    \omega=(x,u,n),
    \qquad
    \omega'=(x',u',n').
\end{equation}
Define the extended-valued base-preserving cost
\begin{equation}
\label{eq:base-preserving-transport-cost}
    c(\omega,\omega')
    =
    \begin{cases}
    d_{\mathcal B}^2
    \bigl((u,n),(u',n')\bigr),
        & x=x',\\[4pt]
    +\infty,
        & x\ne x'.
    \end{cases}
\end{equation}
Let \(\Pi(\mu_t,\nu_t^\theta)\) denote the set of couplings of
\(\mu_t\) and \(\nu_t^\theta\). Then
\begin{equation}
\label{eq:total-space-fiberwise-transport}
    \mathcal W_{\rm fib}^2(t)
    =
    \inf_{\Gamma\in\Pi(\mu_t,\nu_t^\theta)}
    \int
    c(\omega,\omega')\,
    \Gamma(d\omega,d\omega').
\end{equation}
The infinite cost forbids transport between different base points, so
every finite-cost coupling disintegrates into couplings between
\(\mu_t^x\) and \(\nu_t^{\theta,x}\) within the same fiber. Thus the
total-space formulation is exactly equivalent to the base-preserving
fiberwise transport definition of \(R(t)\).

\section{Fixed-Curve Optimization Details}
\label{app:fixed-curve-optimization}

\subsection{Constrained Formulation and KKT Solution}

Let
\begin{equation}
\label{eq:fixed-curve-action-risk-functionals}
    K[w]
    :=
    \int_0^1
    \frac{L'(\tau)^2}{w(\tau)}\,d\tau,
    \qquad
    E_R[w]
    :=
    \int_0^1
    R(\tau)w(\tau)\,d\tau .
\end{equation}
The fixed-curve formulation of
Section~\ref{sec:fixed-curve-model-aware-allocation} is
\begin{equation}
\label{eq:app-fixed-curve-constrained-problem}
    \min_{w>0} K[w]
    \qquad
    \text{subject to}
    \qquad
    E_R[w]\le B,
    \qquad
    \int_0^1 w(\tau)\,d\tau=1 .
\end{equation}
The kinetic term is convex in \(w>0\), while the two constraints are
linear. Under the usual regularity and strict-feasibility conditions, the KKT
conditions characterize the optimum. With multipliers
\(\lambda\ge0\) and \(\eta\), the Lagrangian is
\begin{equation}
\label{eq:allocation-lagrangian}
    \mathcal L[w,\lambda,\eta]
    =
    \int_0^1
    \left[
        \frac{L'(\tau)^2}{w(\tau)}
        +\lambda R(\tau)w(\tau)
        +\eta w(\tau)
    \right]d\tau
    -\lambda B-\eta .
\end{equation}
Pointwise stationarity gives
\begin{equation}
\label{eq:allocation-kkt-stationarity}
    -\frac{L'(\tau)^2}{w(\tau)^2}
    +\lambda R(\tau)+\eta
    =0,
\end{equation}
and hence
\begin{equation}
\label{eq:app-optimal-allocation-density}
    w^\star(\tau)
    =
    \frac{L'(\tau)}
         {\sqrt{\eta+\lambda R(\tau)}} .
\end{equation}
The normalization condition determines \(\eta\), and complementary
slackness gives
\begin{equation}
\label{eq:allocation-complementary-slackness}
    \lambda\bigl(E_R[w^\star]-B\bigr)=0.
\end{equation}
Thus an inactive risk constraint has \(\lambda=0\) and recovers the
kinetic baseline, whereas \(\lambda>0\) implies an active risk constraint
and selects a point on the kinetic--risk tradeoff. Up to the additive
constant \(-\lambda B\), the corresponding penalized objective is the
\(\mathcal J_\lambda\) used in the main text.

\subsection{Discrete Interval Allocation}
\label{app:discrete-interval-allocation}

For a partition \(0=\tau_0<\cdots<\tau_K=1\), define the
interval coefficient-space arc length \(L_k\) and the corresponding
schedule-time allocation \(\Delta t_k\) by
\begin{equation}
\label{eq:interval-length-time-allocation}
    L_k
    :=
    \int_{\tau_k}^{\tau_{k+1}}L'(u)\,du,
    \qquad
    \Delta t_k
    :=
    \Phi(\tau_{k+1})-\Phi(\tau_k),
\end{equation}
and let \(r_k\) denote the interval-average fiberwise risk. The discrete
fixed-curve kinetic and schedule-time-integrated risk terms are
\begin{equation}
\label{eq:discrete-action-risk-functionals}
    K_{\rm disc}
    =
    \sum_k\frac{L_k^2}{\Delta t_k},
    \qquad
    E_{R,{\rm disc}}
    =
    \sum_k r_k\Delta t_k .
\end{equation}
The constrained problem therefore becomes
\begin{equation}
\label{eq:discrete-constrained-allocation}
    \begin{aligned}
    \min_{\Delta t_k>0}\quad&
    \sum_k\frac{L_k^2}{\Delta t_k}\\
    \text{subject to}\quad&
    \sum_k r_k\Delta t_k\le B,
    \qquad
    \sum_k\Delta t_k=1 .
    \end{aligned}
\end{equation}
For fixed \(\lambda\ge0\), the corresponding penalized objective is
\begin{equation}
\label{eq:discrete-penalized-allocation}
    \mathcal J_\lambda^{\rm disc}
    =
    \sum_k\frac{L_k^2}{\Delta t_k}
    +
    \lambda\sum_k r_k\Delta t_k .
\end{equation}
The KKT stationarity condition gives
\begin{equation}
\label{eq:discrete-optimal-allocation}
    \Delta t_k^\star
    =
    \frac{L_k}{\sqrt{\eta+\lambda r_k}},
    \qquad
    \sum_k\Delta t_k^\star=1,
\end{equation}
which is the coefficient-space allocation rule underlying the
one-shot construction in Appendix~\ref{app:model-aware-pipeline}.

\subsection{Effective Geometry and Risk-Recalibration Properties}
\label{app:risk-recalibration-properties}

The kinetic baseline has constant coefficient-space speed.
The continuous model-aware optimum \(w^\star\) admits a similar
interpretation using a risk-modified line element.
Let \(dL=L'(\tau)\,d\tau\) be the coefficient-space arc-length
element. For this solution, define
\begin{equation}
\label{eq:risk-modified-line-element}
    d\widetilde L
    :=
    \frac{dL}{\sqrt{\eta+\lambda R(\tau)}},
    \qquad
    \frac{d\widetilde L}{d\tau}=w^\star(\tau).
\end{equation}
Let \(t^\star\) denote schedule time under the optimal allocation.
Since \(dt^\star=w^\star(\tau)\,d\tau\),
\begin{equation}
\label{eq:risk-modified-constant-speed}
    \frac{d\widetilde L}{dt^\star}=1.
\end{equation}
Thus the optimal allocation gives constant-speed traversal
with respect to \(d\widetilde L\).
The cumulative risk-modified length from the starting point
gives the schedule time \(t^\star\) at each reference location
\(\tau\).
This time assignment acts as an effective model-dependent clock
along the fixed coefficient curve.

The allocation is also invariant to positive affine recalibration of the
risk profile. If
\begin{equation}
\label{eq:affine-risk-recalibration}
    \widetilde R(\tau)=aR(\tau)+b,
    \qquad a>0,
\end{equation}
then choosing
\begin{equation}
\label{eq:recalibrated-multipliers}
    \widetilde\lambda=\frac{\lambda}{a},
    \qquad
    \widetilde\eta
    =
    \eta-\frac{\lambda b}{a}
\end{equation}
gives
\begin{equation}
\label{eq:recalibrated-allocation-denominator}
    \widetilde\eta
    +\widetilde\lambda\widetilde R(\tau)
    =
    \eta+\lambda R(\tau),
\end{equation}
so \(w^\star\) is unchanged. Hence absolute risk scale and offset do not
by themselves determine the deformation.

A related consequence is useful for checkpoint comparisons. For the same parameterized coefficient path and the same \(\lambda>0\), exact equality of two allocation
densities implies
\begin{equation}
\label{eq:equal-allocation-denominators}
    \eta_1+\lambda R_1(\tau)
    =
    \eta_2+\lambda R_2(\tau),
\end{equation}
and therefore
\begin{equation}
\label{eq:equal-allocation-risk-offset}
    R_1(\tau)-R_2(\tau)
    =
    \frac{\eta_2-\eta_1}{\lambda},
\end{equation}
a constant independent of \(\tau\). Thus identical allocations imply
risk-profile agreement up to the additive offset absorbed by the
normalization multiplier.

\section{Model-Aware Construction and Experimental Details}
\label{app:implementation-experimental-details}

\subsection{Model-Aware Schedule Construction and Training Pipeline}
\label{app:model-aware-pipeline}

All model-aware experiments use the same one-shot construction pipeline.
We first train a model under the standard baseline schedule and use a
fixed baseline checkpoint to estimate the fiberwise-risk
profile along the baseline coefficient path. From the estimated risk
profile, we construct the model-aware reparameterization using the
closed-form allocation rule and then train a new model from scratch under
the resulting fixed schedule. The baseline checkpoint is used only for
schedule construction: neither the estimated risk profile nor the schedule
is updated during model-aware training.

Using the partition and interval quantities from
Appendix~\ref{app:discrete-interval-allocation}, let
\(\tau_k^\circ=(\tau_k+\tau_{k+1})/2\) be the midpoint of interval \(k\).
We approximate the interval-average fiberwise risk by the midpoint
Monte Carlo estimate
\begin{equation}
\label{eq:midpoint-risk-estimate}
    r_k
    \approx
    \widehat D_{\rm fib}^2(\tau_k^\circ),
\end{equation}
using the fixed baseline checkpoint and the corresponding fiberwise-risk
specialization from
Section~\ref{sec:general-linear-target-invariance}.
We apply the allocation rule of
Appendix~\ref{app:discrete-interval-allocation} and use the cumulative
interval allocations to define the monotone reparameterization
\(t=\Phi(\tau)\).

Algorithm~\ref{alg:model-aware-pipeline} summarizes the coefficient-space
construction using \(L_k\).
The practical DDPM implementation replaces \(L_k\) with cosine-phase
increments, as described below.
For the straight Cond-OT curve, \(L_k\) is available in closed form.

\begin{algorithm}[t]
\caption{One-shot model-aware schedule construction}
\label{alg:model-aware-pipeline}
\begin{algorithmic}[1]
\Require Baseline coefficient path \(\gamma_0\), baseline checkpoint
with parameters \(\theta_{\rm ref}\), tradeoff \(\lambda\), partition
\(0=\tau_0<\cdots<\tau_K=1\)
\For{\(k=0,\ldots,K-1\)}
    \State Set representative location
    \(\tau_k^\circ=(\tau_k+\tau_{k+1})/2\)
    \State Estimate the midpoint risk
    \(r_k\leftarrow
    \widehat D_{\rm fib}^2(\tau_k^\circ;\theta_{\rm ref})\)
    \State Compute the interval coefficient-space arc length \(L_k\)
\EndFor
\State Solve for \(\eta\) such that
\(\Delta t_k^\star
=L_k/\sqrt{\eta+\lambda r_k}\)
and \(\sum_k\Delta t_k^\star=1\)
\State Construct \(t=\Phi(\tau)\) from the cumulative
\(\{\Delta t_k^\star\}_{k=0}^{K-1}\)
\State Realize \(\Phi\) in the model-family-specific training schedule
\State Train a new model from scratch with \(\Phi\) fixed
\end{algorithmic}
\end{algorithm}

Here and below, \(\widehat{\E}\) denotes the empirical average over the
sampled endpoint pairs at the indicated reference location. Numerical
risk estimates additionally average squared residuals over all channels
and spatial positions; all reported \(\lambda\) values use this
per-coordinate convention.

\paragraph{DDPM.}
For DDPM, we retain the standard offset-normalized cosine schedule of
Appendix~\ref{app:vp-kinetic-reference-coordinate}.
Its baseline time \(\tau\) is the practical VP reference coordinate.
With \(\varphi(\tau)\) denoting the cosine phase,
\begin{equation}
\label{eq:ddpm-baseline-coefficient-path}
    \bar\alpha(\tau)
    =
    \frac{\cos^2\varphi(\tau)}
         {\cos^2\varphi(0)},
    \qquad
    \gamma_0(\tau)
    =
    \left(
        \sqrt{\bar\alpha(\tau)},
        \sqrt{1-\bar\alpha(\tau)}
    \right).
\end{equation}
On this unit-radius VP coefficient curve, the interval coefficient-space
arc length is
\begin{equation}
\label{eq:vp-interval-coefficient-arc-length}
    L_k
    =
    \arccos\sqrt{\bar\alpha(\tau_{k+1})}
    -
    \arccos\sqrt{\bar\alpha(\tau_k)}.
\end{equation}
In the practical DDPM implementation, we replace \(L_k\) with the
cosine-phase increment
\begin{equation}
\label{eq:cosine-phase-increment}
    \Delta\varphi_k
    :=
    \varphi(\tau_{k+1})-\varphi(\tau_k)
\end{equation}
and use
\begin{equation}
\label{eq:practical-ddpm-interval-allocation}
    \Delta t_k^\star
    =
    \frac{\Delta\varphi_k}
         {\sqrt{\eta+\lambda r_k}},
    \qquad
    \sum_k\Delta t_k^\star=1.
\end{equation}
Because \(\varphi(\tau)\) is affine in \(\tau\), setting \(\lambda=0\)
gives \(\Delta t_k^\star=\tau_{k+1}-\tau_k\) and thus recovers the
standard cosine baseline timing.
Nonconstant fiberwise risk deforms the allocation over the practical
VP reference coordinate.

For this continuous schedule, the replacement has a small total
approximation error in coefficient-space arc length.
For \(\varphi(0)<\varphi<\pi/2\),
\begin{equation}
\label{eq:vp-angle-phase-derivative}
    \frac{d}{d\varphi}
    \arccos\frac{\cos\varphi}{\cos\varphi(0)}
    =
    \frac{\sin\varphi}
         {\sqrt{\sin^2\varphi-\sin^2\varphi(0)}}
    \ge 1.
\end{equation}
Hence \(L_k\ge\Delta\varphi_k\) on every interval.
Since the curve has total arc length \(\pi/2\), the total absolute
approximation error relative to this length is
\begin{equation}
\label{eq:aggregate-arc-phase-discrepancy}
    \frac{\sum_k|L_k-\Delta\varphi_k|}{\sum_k L_k}
    =
    1-\frac{\varphi(1)-\varphi(0)}{\pi/2}
    =
    \frac{s_{\rm off}}{1+s_{\rm off}}
    \approx 0.79\%
\end{equation}
for \(s_{\rm off}=0.008\).

We estimate the risk on 199 reference intervals.
For the primary \(\epsilon\)-prediction setting, the midpoint risk is
\begin{equation}
\label{eq:ddpm-noise-risk-estimate}
    r_k
    =
    \bigl(1-\bar\alpha_{\tau_k^\circ}\bigr)
    \widehat{\E}
    \left[
        \left\|
        \epsilon_\theta(x_{\tau_k^\circ},\tau_k^\circ)
        -\epsilon
        \right\|^2
    \right].
\end{equation}

For the \(v\)-prediction robustness setting
\citep{salimans2022progressive}, let \(t_{\rm zsnr}(\tau)\) denote
the interpolated model-time coordinate under the
zero-terminal-SNR-rescaled scheduler, defined by
\begin{equation}
\label{eq:zero-snr-reference-time-map}
    \bar\alpha^{\rm zsnr}
    \bigl(t_{\rm zsnr}(\tau)\bigr)
    =
    \bar\alpha(\tau).
\end{equation}
With
\begin{equation}
\label{eq:ddpm-v-state-target}
    \begin{aligned}
    x_\tau
    &=
    \sqrt{\bar\alpha(\tau)}\,x_0
    +
    \sqrt{1-\bar\alpha(\tau)}\,\epsilon,\\
    v_\tau
    &=
    \sqrt{\bar\alpha(\tau)}\,\epsilon
    -
    \sqrt{1-\bar\alpha(\tau)}\,x_0,
    \end{aligned}
\end{equation}
the midpoint risk is
\begin{equation}
\label{eq:ddpm-v-risk-estimate}
    r_k
    =
    \bar\alpha(\tau_k^\circ)
    \bigl(1-\bar\alpha(\tau_k^\circ)\bigr)
    \widehat{\E}
    \left[
        \left\|
        v_\theta
        \bigl(
            x_{\tau_k^\circ},
            t_{\rm zsnr}(\tau_k^\circ)
        \bigr)
        -
        v_{\tau_k^\circ}
        \right\|^2
    \right].
\end{equation}

After constructing the model-aware traversal, we realize it on the
native \(T=1000\) diffusion grid. For model-time boundaries
\(t_j=j/T\), let
\begin{equation}
\label{eq:discrete-reference-phase-map}
    \tau_j=\Phi^{-1}(t_j),
    \qquad
    \varphi_j=\varphi(\tau_j).
\end{equation}
The discrete diffusion coefficients are obtained directly from
consecutive phase values:
\begin{equation}
\label{eq:discrete-diffusion-coefficients}
    \alpha_j
    =
    \frac{\cos^2\varphi_j}
         {\cos^2\varphi_{j-1}},
    \qquad
    \beta_j=1-\alpha_j,
    \qquad
    j=1,\ldots,T,
\end{equation}
with the standard \(0.999\) upper cap on \(\beta_j\).

Both the baseline and model-aware \(v\)-prediction runs additionally apply
the same zero-terminal-SNR rescaling before the final \(\beta_j\) cap. For a discrete diffusion
schedule, let \(q_j=\sqrt{\bar\alpha_j}\). We rescale
\begin{equation}
\label{eq:zero-terminal-snr-rescaling}
    q_j^{\rm zsnr}
    =
    q_{\rm first}
    \frac{q_j-q_{\rm last}}
         {q_{\rm first}-q_{\rm last}},
    \qquad
    \bar\alpha_j^{\rm zsnr}
    =
    \bigl(q_j^{\rm zsnr}\bigr)^2 .
\end{equation}
The per-step coefficients are reconstructed from the rescaled cumulative
coefficients, with the resulting \(\beta_j\) values capped at \(0.999\)
for numerical stability. Because this cap is applied after rescaling, the
realized terminal SNR is small but nonzero: \(2.278\times10^{-9}\) for the
baseline and \(8.512\times10^{-10}\) for the model-aware schedule with
\(\lambda=220\). For \(v\)-prediction evaluation, both baseline
and model-aware runs use trailing timestep spacing
\citep{lin2024flawed}.

In both prediction settings, training uses the resulting fixed diffusion
coefficients, with model-time indices sampled uniformly as in the
corresponding baseline. Thus the model-aware construction changes the
diffusion schedule itself rather than the distribution used to sample
training timesteps. Each model-aware network is independently initialized
and trained from scratch.

\paragraph{Flow matching.}
For flow matching, we fix the Cond-OT coefficient curve
\begin{equation}
\label{eq:fm-baseline-coefficient-path}
    \gamma_0(\tau)=(1-\tau,\tau),
    \qquad
    x_\tau=(1-\tau)x_0+\tau\epsilon ,
\end{equation}
and estimate the fiberwise risk on 200 reference intervals. For velocity
prediction \(v_\tau=\epsilon-x_0\), the midpoint risk is
\begin{equation}
\label{eq:fm-velocity-risk-estimate}
    r_k
    =
    \tau_k^{\circ\,2}
    (1-\tau_k^\circ)^2
    \widehat{\E}
    \left[
        \left\|
        v_\theta(x_{\tau_k^\circ},\tau_k^\circ)
        -(\epsilon-x_0)
        \right\|^2
    \right].
\end{equation}
For the straight Cond-OT coefficient curve, the interval
coefficient-space arc length is
\begin{equation}
\label{eq:cond-ot-interval-coefficient-arc-length}
    L_k
    =
    \sqrt{2}\,(\tau_{k+1}-\tau_k).
\end{equation}
We use the normalized Euclidean coefficient-space convention of the
main text. On the fixed Cond-OT curve, the pathwise action is
proportional to \(\mathcal J_{\rm kin}\); this constant factor is
absorbed into the kinetic--risk tradeoff convention
(Appendix~\ref{app:pathwise-action-coefficient-kinetics}).

The resulting interval allocations define a piecewise-linear
reparameterization \(t=\Phi(\tau)\). During model-aware training, model
time \(t\) is drawn from the same training-time sampling distribution as in the
corresponding baseline and mapped to
\begin{equation}
\label{eq:fm-inverse-time-map}
    \tau=\Phi^{-1}(t).
\end{equation}
Holding the training-time sampling distribution fixed in model time
\(t\) generally induces a different reference-coordinate distribution
over \(\tau\).
The state and velocity target are then
\begin{equation}
\label{eq:fm-training-state-target}
    x_t=(1-\tau)x_0+\tau\epsilon,
    \qquad
    \frac{dx_t}{dt}
    =
    \frac{\epsilon-x_0}{w(\tau)},
    \qquad
    w(\tau)=\Phi'(\tau).
\end{equation}
Hence reparameterization changes both the state--time map and the
corresponding velocity target. Within every baseline--model-aware
comparison, the endpoint coupling and training-time sampling distribution
are held fixed; only the traversal of the prescribed coefficient curve is
changed.

The Monte Carlo budgets and computational cost of the one-time
risk-profile estimation are reported in
Appendix~\ref{app:risk-profile-checkpoint-sensitivity}.

\subsection{Architectures, Training, and Evaluation Details}
\label{app:architectures-training-evaluation}

\paragraph{Model architectures.}
The primary experiments use U-Net backbones \citep{ronneberger2015unet}. For CIFAR-10 DDPM, we use a
Diffusers-style unconditional U-Net \citep{vonplaten2022diffusers} with four resolution levels, channel
widths \((128,128,256,256)\), and two layers per block. The architecture-family
control replaces this backbone with the CIFAR-10 U-ViT-S/2 architecture:
\(32\times32\) inputs, patch size \(2\), embedding dimension \(512\),
depth \(12\), and \(8\) attention heads. It is unconditional and uses the
same \(\epsilon\)-prediction target \citep{bao2023uvit}.
For CIFAR-10 flow matching, we follow the TorchCFM CIFAR-10
image-generation configuration, with \(128\) base channels, two residual
blocks per resolution, channel multipliers \((1,2,2,2)\), and four-head
attention at \(16\times16\) resolution \citep{tong2024improving}.

For ImageNet-64, DDPM and flow matching use the same unconditional
U-Net architecture at \(64\times64\) resolution, with channel widths
\((192,384,384,768)\) and two layers per block. Attention is used at
\(32\times32\), \(16\times16\), and \(8\times8\) resolutions, with
attention-head dimension \(8\); we use \(32\) GroupNorm groups and no
dropout.

\paragraph{Optimization.}
The CIFAR-10 DDPM U-Net and U-ViT runs use the same optimization recipe:
AdamW with batch size \(128\), learning rate \(2\times10^{-4}\), weight decay \(10^{-4}\), and an exponential moving average (EMA) with
decay \(0.9999\). The CIFAR-10 flow-matching
models use Adam with batch size \(128\), learning rate \(2\times10^{-4}\),
zero weight decay, \(5000\) warmup steps, gradient clipping at \(1.0\),
and EMA decay \(0.9999\).

On ImageNet-64, both model families use AdamW with learning rate
\(10^{-4}\), zero weight decay, EMA decay \(0.9999\), effective batch
size \(128\), and mixed-precision training. The DDPM runs use two GPUs with per-GPU batch size \(32\) and two
gradient-accumulation steps. The flow-matching runs instead use two GPUs
with per-GPU batch size \(64\) and no gradient accumulation, so each
minibatch-OT coupling is formed within a \(64\)-sample per-GPU batch.

\paragraph{Evaluation.}
All reported FIDs use \(50{,}000\) generated samples. FID differences, relative reductions, and gain-recovery percentages are
computed from unrounded FIDs. Within each
baseline--model-aware comparison, we match the evaluation protocol,
generated-sample seed, sampler or integrator, and NFE budget. Unless otherwise specified, single-seed FID evaluations use DPM++3M
for DDPM and midpoint integration for flow matching, both at 16 NFE. For
CIFAR-10, FID is computed against the full training split. For
ImageNet-64, we use the training split of the Downsampled ImageNet
\(64\times64\) dataset distributed through Academic Torrents and compute
FID against its full validation split. Exact epochs or update counts are
stated with the corresponding results. For CIFAR-10, one epoch
corresponds to \(390\) optimizer updates.

\subsection{Pretrained Conditional Latent-Model Diagnostics}
\label{app:pretrained-latent-diagnostics}

\paragraph{Conditional latent-state extension.}
In this subsection, \(x\) denotes the checkpoint-native latent state rather
than a pixel-space observation, and \(c\) denotes a class label or text
prompt. We allow an arbitrary joint endpoint law
\begin{equation}
\label{eq:conditional-latent-endpoint-law}
    (x_0,\epsilon,c)\sim\rho,
    \qquad
    \epsilon\sim\mathcal N(0,I),
\end{equation}
and average the fiberwise risk over this joint law. For a conditional
predictor, we regard \((x,c)\) as the fixed base variable, replace the
conditional fiber laws \(\mu_t^x\) and \(\nu_t^{\theta,x}\) by
\(\mu_t^{x,c}\) and \(\nu_t^{\theta,x,c}\), and otherwise retain the same
decomposition fiber \(\mathcal F_x\). The identities of
Section~\ref{sec:fiberwise-prediction-risk} therefore remain unchanged.
Each checkpoint is evaluated in its own native latent representation; no
coordinate-wise alignment between the DiT and 2-RF latent spaces is assumed.

\paragraph{DiT-XL/2.}
We use the public \texttt{facebook/DiT-XL-2-256} checkpoint, a 675M
class-conditional latent diffusion transformer on ImageNet-256
\citep{peebles2023dit}.
We select 64 examples from a fixed seed-42 shuffle of an ImageNet-1K
validation subset and apply the DiT/ADM center crop at \(256\times256\)
\citep{deng2009imagenet,dhariwal2021adm}. We then draw one sample per
example from the checkpoint VAE posterior using its native scaling factor. Each latent is paired with independent Gaussian noise and its class
label. We query 100 distinct diffusion timesteps \(t(\tau)\) nearest to uniform
midpoints in the same practical VP reference coordinate used in the
other diffusion comparisons. The checkpoint is evaluated without
classifier-free guidance; from its eight output channels we retain only the
four \(\epsilon\)-prediction channels and exclude the learned-variance
channels. With the squared error averaged over latent coordinates, the risk
is
\begin{equation}
\label{eq:dit-fiberwise-risk}
    R_{\rm DiT}(\tau)
    =
    \bigl(1-\bar\alpha_{t(\tau)}\bigr)
    \E\left\|
        \epsilon_\theta(x_\tau,t(\tau),c)-\epsilon
    \right\|^2 .
\end{equation}

\paragraph{InstaFlow 2-RF.}
We use the public pre-distillation checkpoint
\texttt{XCLiu/2\_rectified\_flow\_from\_sd\_1\_5}, a 0.9B
text-conditional latent 2-Rectified Flow model
\citep{liu2023flow,liu2024instaflow}.
To construct a public model-induced coupling, we select 64 nonempty prompts
from a fixed seed-42 shuffle of the COCO 2017 training-caption stream and
sample one Gaussian noise latent per prompt. Using the same prompt and noise,
the Stable Diffusion 1.5 teacher produces the paired endpoint with a
second-order DPM-Solver++ sampler at 25 steps and guidance scale \(5.0\).
Both endpoints remain in the native Stable Diffusion latent scale, without
VAE decoding or re-encoding. This defines a reconstructed reflow coupling induced by the teacher
under a public COCO prompt distribution. It differs from the original
InstaFlow training coupling and the minibatch OT endpoint coupling
used in our OT-CFM experiments
\citep{lin2014coco,rombach2022latent,lu2025dpmsolverpp,ho2022classifierfree}.

We evaluate the released 2-RF predictor with its own text encoder and the raw
text condition, without classifier-free guidance. In the common data-to-noise
coordinate,
\begin{equation}
\label{eq:two-rf-data-noise-coordinate}
    x_\tau=(1-\tau)x_0+\tau\epsilon,
    \qquad
    \tau=0\ \text{at the teacher endpoint},
    \quad
    \tau=1\ \text{at noise}.
\end{equation}
The official model timestep is \(1000\tau\): it equals \(1000\) at the noise
endpoint and decreases toward \(0\) along the native noise-to-data sampling
direction. The released predictor uses the corresponding native velocity
orientation \(x_0-\epsilon\), so the data-to-noise fiberwise risk can be
evaluated equivalently as
\begin{equation}
\label{eq:two-rf-fiberwise-risk}
    R_{\rm 2RF}(\tau)
    =
    \tau^2(1-\tau)^2
    \E\left\|
        v_\theta(x_\tau,1000\tau,c)-(x_0-\epsilon)
    \right\|^2 .
\end{equation}
Flipping both the prediction and target gives the identical squared residual
in the data-to-noise velocity orientation.

\paragraph{Estimation and allocation comparison.}
Each profile uses the same 64 fixed endpoint pairs at all 100 reference
locations. Nested subsample comparisons at \(N\in\{16,32,64\}\) show that
the reported shape metrics have stabilized by \(N=64\). Per-sample squared
errors are averaged over latent coordinates, and the resulting profiles are
interpolated to the common midpoint grid and compared with the same
normalized-risk and allocation-density metrics as in
Appendix~\ref{app:shared-risk-allocation-agreement}. Allocation densities are
constructed from the raw risks on a 10,000-point midpoint grid, using the
previously fixed family values \(\lambda=220\) for DiT and \(\lambda=450\)
for 2-RF. For DiT, we additionally apply the affine risk-recalibration
property of Appendix~\ref{app:risk-recalibration-properties} to diagnose the
observed absolute-scale shift.

\section{Additional Experimental Results and Diagnostics}
\label{app:additional-results-diagnostics}

\subsection{Complete CIFAR-10 DDPM Sampler--NFE Grid}
\label{app:ddpm-cifar-full-grid}

Table~\ref{tab:ddpm-cifar-full-grid} reports the complete sampler--NFE grid
underlying the main CIFAR-10 DDPM \(\epsilon\)-prediction results. The main
text uses DPM++3M as the primary sampler at 16, 32, and 64 NFE and reports
DDIM and DPM++2M at 16 NFE as cross-sampler checks. Here we include the
remaining DDIM and DPM++2M budgets for completeness
\citep{song2021ddim,lu2025dpmsolverpp}. The model-aware schedule improves for all three paired seeds in every one
of the nine configurations, with mean relative FID reductions ranging from
\(9.2\%\) to \(16.4\%\).

\begin{table}[t]
    \centering
    \small
    \setlength{\tabcolsep}{6pt}
    \caption{
        CIFAR-10 DDPM FID on the complete sampler--NFE grid with
        \(\epsilon\)-prediction and \(\lambda=220\). Values are mean
        \(\pm\) standard deviation over three paired training seeds.
    }
    \label{tab:ddpm-cifar-full-grid}
    \begin{tabular}{lccccc}
    \toprule
    & &
    \multicolumn{1}{c}{\textbf{Baseline}} &
    \multicolumn{3}{c}{\textbf{Model-aware (Ours)}} \\
    \cmidrule(lr){3-3}
    \cmidrule(lr){4-6}

    Sampler
    & NFE
    & FID~\(\downarrow\)
    & FID~\(\downarrow\)
    & \(\Delta\)FID~\(\uparrow\)
    & \%Impr.~\(\uparrow\) \\
    \midrule
    
        DDIM
        & 16
        & \(11.27 \pm 0.19\)
        & \(9.98 \pm 0.23\)
        & \(1.29 \pm 0.20\)
        & \(11.4 \pm 1.7\%\) \\
        DDIM
        & 32
        & \(8.14 \pm 0.23\)
        & \(7.39 \pm 0.23\)
        & \(0.75 \pm 0.14\)
        & \(9.2 \pm 1.7\%\) \\
        DDIM
        & 64
        & \(7.27 \pm 0.23\)
        & \(6.44 \pm 0.21\)
        & \(0.83 \pm 0.07\)
        & \(11.4 \pm 0.8\%\) \\
        \midrule
        DPM++2M
        & 16
        & \(9.81 \pm 0.26\)
        & \(8.29 \pm 0.14\)
        & \(1.51 \pm 0.14\)
        & \(15.4 \pm 1.1\%\) \\
        DPM++2M
        & 32
        & \(7.90 \pm 0.27\)
        & \(7.00 \pm 0.25\)
        & \(0.90 \pm 0.12\)
        & \(11.4 \pm 1.5\%\) \\
        DPM++2M
        & 64
        & \(6.97 \pm 0.24\)
        & \(6.30 \pm 0.27\)
        & \(0.67 \pm 0.07\)
        & \(9.7 \pm 1.1\%\) \\
        \midrule
        DPM++3M
        & 16
        & \(9.64 \pm 0.26\)
        & \(8.06 \pm 0.16\)
        & \(1.58 \pm 0.11\)
        & \(\mathbf{16.4 \pm 0.7\%}\) \\
        DPM++3M
        & 32
        & \(7.69 \pm 0.26\)
        & \(6.79 \pm 0.27\)
        & \(0.91 \pm 0.12\)
        & \(11.8 \pm 1.5\%\) \\
        DPM++3M
        & 64
        & \(6.83 \pm 0.24\)
        & \(6.19 \pm 0.27\)
        & \(0.64 \pm 0.07\)
        & \(9.3 \pm 1.2\%\) \\
        \bottomrule
    \end{tabular}
\end{table}

\subsection{\texorpdfstring{Linear-\(\beta\) DDPM Control}
{Linear-beta DDPM Control}}
\label{app:linear-beta-ddpm-control}

Table~\ref{tab:ddpm-linear-schedule} reports results for the linear-\(\beta\) DDPM control
used in Section~\ref{sec:ddpm-cifar-main}. The control uses \(T=1000\)
diffusion steps with \(\beta_t\) linearly spaced from \(10^{-4}\) to
\(2\times10^{-2}\). Architecture, optimizer, batch size, training horizon,
data pipeline, and all other training settings are identical to the
cosine-DDPM baseline; only the diffusion schedule is changed. All results use
epoch-400 EMA weights.

\begin{table}[t]
    \centering
    \small
    \setlength{\tabcolsep}{12pt}
    \caption{
        CIFAR-10 DDPM FID for the cosine and linear schedules.
        Single-seed results.
    }
    \label{tab:ddpm-linear-schedule}
    \begin{tabular}{lccc}
        \toprule
        Sampler
        & NFE
        & FID (Cosine)~\(\downarrow\)
        & FID (Linear)~\(\downarrow\) \\
        \midrule
        DDIM     & 16 & \(11.08\) & \(13.77\) \\
        DDIM     & 32 & \(7.96\)  & \(10.30\) \\
        DDIM     & 64 & \(7.05\)  & \(8.66\) \\
        \midrule
        DPM++2M  & 16 & \(9.60\)  & \(13.93\) \\
        DPM++2M  & 32 & \(7.73\)  & \(11.34\) \\
        DPM++2M  & 64 & \(6.78\)  & \(9.59\) \\
        \midrule
        DPM++3M  & 16 & \(9.43\)  & \(13.54\) \\
        DPM++3M  & 32 & \(7.52\)  & \(11.10\) \\
        DPM++3M  & 64 & \(6.64\)  & \(9.36\) \\
        \bottomrule
    \end{tabular}
\end{table}

The linear schedule is consistently weaker than cosine across all nine
sampler--NFE configurations, supporting our use of cosine as the primary
DDPM baseline in the main text.

\subsection{\texorpdfstring{DDPM \(v\)-Prediction Results}
{DDPM v-Prediction Results}}
\label{app:ddpm-v-prediction-results}

We additionally evaluate the model-aware schedule under \(v\)-prediction
to test prediction-target robustness within DDPM and provide an additional
instance of the general linear-target formulation. Both baseline and
model-aware runs use matched zero-terminal-SNR rescaling, and evaluation
uses trailing timestep spacing. We retain the same \(\lambda=220\) used
for \(\epsilon\)-prediction. We report DPM-Solver++ multistep results. In the Hugging Face Diffusers
\texttt{DDIMScheduler} implementation used in our experiments
\citep{vonplaten2022diffusers}, DDIM updates with trailing timestep
spacing do not always end at the next grid timestep at these NFE budgets.
We therefore omit DDIM from this robustness table.

\begin{table}[t]
    \centering
    \small
    \setlength{\tabcolsep}{12pt}
    \caption{
        CIFAR-10 DDPM \(v\)-prediction FID at epoch 400 with \(\lambda=220\)
        under matched zero-terminal-SNR rescaling and trailing timestep spacing.
        Single-seed results.
    }
    \label{tab:ddpm-v-prediction-results}
    \begin{tabular}{lccccc}
    
    \toprule
    & &
    \multicolumn{1}{c}{\textbf{Baseline}} &
    \multicolumn{3}{c}{\textbf{Model-aware (Ours)}} \\
    \cmidrule(lr){3-3}
    \cmidrule(lr){4-6}

    Sampler
    & NFE
    & FID~\(\downarrow\)
    & FID~\(\downarrow\)
    & \(\Delta\)FID~\(\uparrow\)
    & \%Impr.~\(\uparrow\) \\
    \midrule
    
        DPM++2M  & 16 & \(11.60\) & \(9.94\) & \(1.66\) & \(14.3\%\) \\
        DPM++2M  & 32 & \(9.24\)  & \(8.27\) & \(0.97\) & \(10.5\%\) \\
        DPM++2M  & 64 & \(8.03\)  & \(7.40\) & \(0.63\) & \(7.9\%\) \\
        \midrule
        DPM++3M  & 16 & \(11.31\) & \(9.65\) & \(1.66\) & \(\mathbf{14.7\%}\) \\
        DPM++3M  & 32 & \(8.98\)  & \(8.02\) & \(0.96\) & \(10.7\%\) \\
        DPM++3M  & 64 & \(7.85\)  & \(7.27\) & \(0.58\) & \(7.4\%\) \\
        \bottomrule
    \end{tabular}
\end{table}

Table~\ref{tab:ddpm-v-prediction-results} shows that the model-aware
schedule improves all six reported DPM-Solver++ configurations across
both second- and third-order multistep solvers, with relative FID
reductions from \(7.4\%\) to \(14.7\%\). The gains are therefore
preserved under a different prediction target.

\subsection{CIFAR-10 U-ViT Architecture-Family Transfer}
\label{app:ddpm-uvit-results}

We evaluate architecture-family transfer by replacing the CIFAR-10 DDPM
U-Net with U-ViT-S/2. We construct the model-aware schedule from the U-ViT baseline risk
profile with the same \(\lambda=220\), without architecture-specific
retuning, and then hold it fixed. Both baseline and
model-aware FIDs are evaluated at 600 training epochs using the same
sampler--NFE configurations as the main CIFAR-10 DDPM results.

\begin{table}[t]
    \centering
    \small
    \setlength{\tabcolsep}{12pt}
    \caption{
        CIFAR-10 U-ViT-S/2 architecture-transfer FID at 600 epochs
        with \(\lambda=220\). Single-seed results.
    }
    \label{tab:ddpm-uvit-results}
    \begin{tabular}{lccccc}
    
    \toprule
    & &
    \multicolumn{1}{c}{\textbf{Baseline}} &
    \multicolumn{3}{c}{\textbf{Model-aware (Ours)}} \\
    \cmidrule(lr){3-3}
    \cmidrule(lr){4-6}

    Sampler
    & NFE
    & FID~\(\downarrow\)
    & FID~\(\downarrow\)
    & \(\Delta\)FID~\(\uparrow\)
    & \%Impr.~\(\uparrow\) \\
    \midrule
    
        DDIM
        & 16
        & \(15.38\)
        & \(14.17\)
        & \(1.21\)
        & \(7.8\%\) \\
        DPM++2M
        & 16
        & \(13.50\)
        & \(12.05\)
        & \(1.46\)
        & \(10.8\%\) \\
        DPM++3M
        & 16
        & \(13.18\)
        & \(11.57\)
        & \(1.61\)
        & \(\mathbf{12.2\%}\) \\
        DPM++3M
        & 32
        & \(10.52\)
        & \(9.70\)
        & \(0.83\)
        & \(7.9\%\) \\
        DPM++3M
        & 64
        & \(9.17\)
        & \(8.69\)
        & \(0.48\)
        & \(5.2\%\) \\
        \bottomrule
    \end{tabular}
\end{table}

Table~\ref{tab:ddpm-uvit-results} shows improvements in all five reported
sampler--NFE configurations, including a \(12.2\%\) relative FID reduction with
DPM++3M at 16 NFE. Thus the model-aware improvement persists after
changing the architecture family without architecture-specific retuning.

\subsection{Flow-Matching Coupling and Training-Time Sampling Robustness}
\label{app:fm-coupling-time-sampling}

We test whether the model-aware improvement persists under changes to
endpoint coupling and the training-time sampling distribution. Within
each matched 400-epoch setting, the baseline and model-aware runs use the
same coupling and training-time sampling distribution. We retain
\(\lambda=450\) throughout without retuning as a robustness test.

The logit-normal control uses \(t=\operatorname{sigmoid}(z)\) with
\(z\sim\mathcal N(0,1)\), whereas the RF++ U-shaped control uses
\(p(t)\propto e^{4t}+e^{4(1-t)}\) on \(t\in[0,1]\). No additional
timestep clipping is applied during training
\citep{esser2024scaling,lee2024improving}.

\begin{table}[t]
    \centering
    \small
    \setlength{\tabcolsep}{12pt}
    \caption{
        CIFAR-10 flow-matching FID under changes in endpoint coupling and
        the training-time sampling distribution, using matched 400-epoch
        settings. Single-seed results.
    }
    \label{tab:fm-coupling-time-sampling}
    \begin{tabular}{llccc}
    \toprule
    & &
    \multicolumn{1}{c}{\textbf{Baseline}} &
    \multicolumn{2}{c}{\textbf{Model-aware (Ours)}}
    \\
    \cmidrule(lr){3-3}
    \cmidrule(lr){4-5}

    Variation
    & Setting
    & FID~\(\downarrow\)
    & FID~\(\downarrow\)
    & \%Impr.~\(\uparrow\) \\
    \midrule
    
        Reference
        & OT-CFM
        & \(8.60\)
        & \(5.57\)
        & \(35.2\%\) \\
        Coupling
        & Independent-CFM
        & \(8.82\)
        & \(6.06\)
        & \(31.3\%\) \\
        Time sampling
        & Logit-normal
        & \(8.27\)
        & \(5.01\)
        & \(\mathbf{39.4\%}\) \\
        Time sampling
        & U-shaped (RF++)
        & \(8.97\)
        & \(6.12\)
        & \(31.8\%\) \\
        \bottomrule
    \end{tabular}
\end{table}

Table~\ref{tab:fm-coupling-time-sampling} shows improvements in all four
settings, with relative FID reductions from \(31.3\%\) to \(39.4\%\).
The gains persist under independent endpoint coupling and alternative
training-time sampling distributions, while preserving the baseline
ordering across the tested settings.

\subsection{Training-Stage and Tradeoff-Weight Sensitivity}
\label{app:training-stage-tradeoff-sensitivity}

We selected the operating values using preliminary single-seed,
10,000-sample FID sweeps in the two primary CIFAR-10 settings,
before the cross-system risk--allocation analysis.
For DDPM, the sweep used 16-step DDIM at epoch 320 and considered
\(\lambda\in\{155,195,210,220,230,250\}\).
For flow matching, the sweep used 16-NFE midpoint integration
at epoch 400 and considered
\(\lambda\in\{100,420,430,450,460,470,480,490,500\}\).
Among the tested values, \(\lambda=220\) and \(\lambda=450\)
gave the lowest FID for DDPM and flow matching, respectively.
We report representative alternative nonzero values below
to assess sensitivity to the tradeoff weight.

\begin{figure*}[t]
    \centering
    \includegraphics[width=0.48\textwidth]{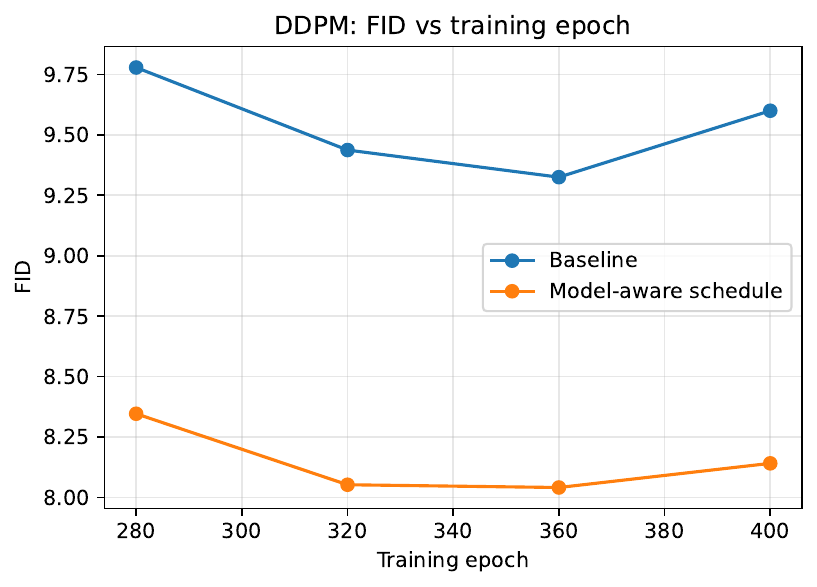}
    \hfill
    \includegraphics[width=0.48\textwidth]{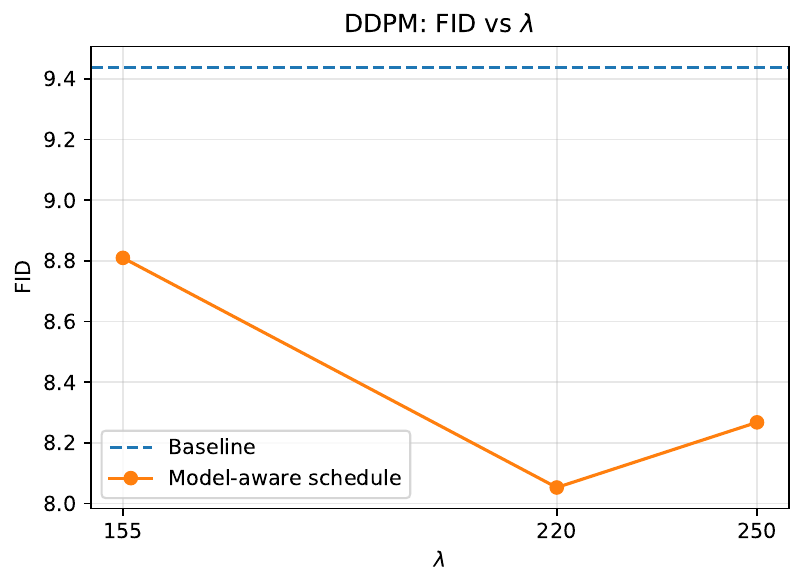}

    \vspace{0.6em}

    \includegraphics[width=0.48\textwidth]{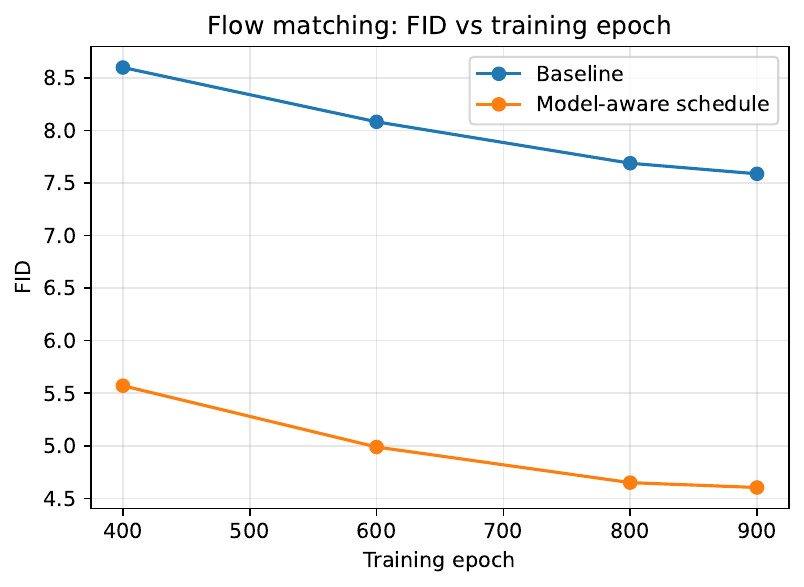}
    \hfill
    \includegraphics[width=0.48\textwidth]{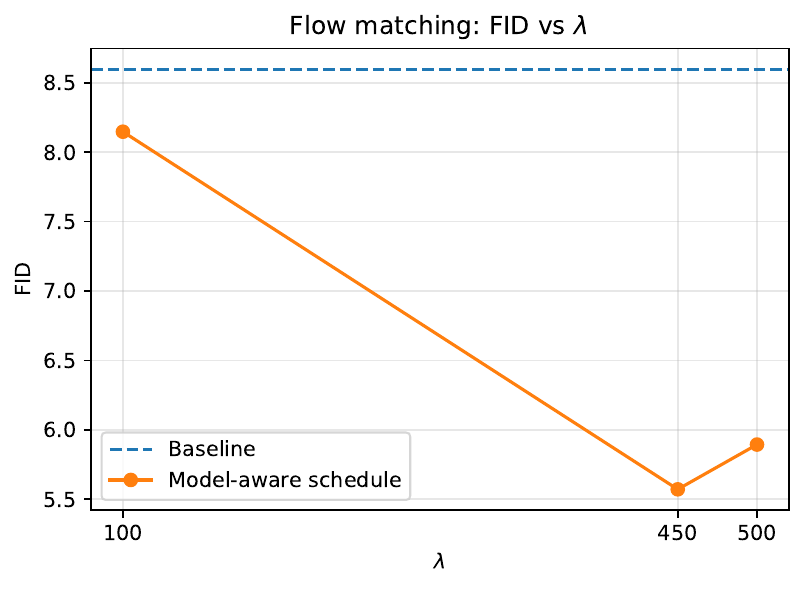}
    \caption{
        FID sensitivity to training stage and \(\lambda\) on CIFAR-10.
        From top left to bottom right: DDPM training stage, DDPM \(\lambda\),
        flow-matching training stage, and flow-matching \(\lambda\).
        DDPM uses \(\epsilon\)-prediction with DPM++3M at 16 NFE; flow
        matching uses midpoint at 16 NFE. The \(\lambda\) panels show only
        tested nonzero values; dashed lines denote the corresponding
        baselines, and connecting lines are visual guides.
    }
    \label{fig:training-stage-tradeoff-sensitivity}
\end{figure*}

We further examine how generation quality varies across model-training
checkpoints and across choices of the tradeoff weight \(\lambda\). Figure~\ref{fig:training-stage-tradeoff-sensitivity} summarizes the corresponding
CIFAR-10 results for both DDPM and flow matching.

For DDPM \(\epsilon\)-prediction, the model-aware schedule consistently
improves over the cosine baseline from epochs 280 through 400. At epoch 320, all tested nonzero values
\(\lambda\in\{155,220,250\}\) also improve over cosine, although the
magnitude of the gain varies across the tested weights.

For flow matching, the model-aware schedule yields an approximately
\(3\)-FID improvement throughout the tested range from epochs 400 to 900. At epoch 400, all
tested nonzero values \(\lambda\in\{100,450,500\}\) improve over the
standard Cond-OT parameterization. Together, these results show that the
observed gains are not tied to a single training stage or an isolated
nonzero choice of \(\lambda\).

\subsection{Risk-Profile Checkpoint Sensitivity}
\label{app:risk-profile-checkpoint-sensitivity}

Tables~\ref{tab:ddpm-risk-checkpoint-sensitivity} and~\ref{tab:fm-risk-checkpoint-sensitivity} give the complete
results for the early-checkpoint studies summarized in the main text.
In both model families, using the risk profile estimated at epoch 100
retains essentially the full improvement obtained with the later estimate, changing the final FID by
only \(0.02\) for DDPM and \(0.06\) for flow matching.

\begin{table}[t]
    \centering
    \small
    \setlength{\tabcolsep}{12pt}
    \caption{
        DDPM \(\epsilon\)-prediction FID sensitivity to the checkpoint used to
        estimate the risk profile. FIDs are evaluated at epoch 400. Single-seed results.
    }
    \label{tab:ddpm-risk-checkpoint-sensitivity}
    \begin{tabular}{lcc}
        \toprule
        Schedule & FID~\(\downarrow\) & \(\Delta\)FID~\(\uparrow\) \\
        \midrule
        Cosine baseline                 & \(9.43\) & -- \\
        Model-aware, epoch 320 estimate & \(7.90\) & \(1.52\) \\
        Model-aware, epoch 100 estimate & \(7.88\) & \(1.55\) \\
        \bottomrule
    \end{tabular}
\end{table}

\begin{table}[t]
    \centering
    \small
    \setlength{\tabcolsep}{12pt}
    \caption{
        Flow-matching FID sensitivity to the checkpoint used to estimate the
        risk profile in the 400-epoch reference OT-CFM setting.
        Single-seed results.
    }
    \label{tab:fm-risk-checkpoint-sensitivity}
    \begin{tabular}{lcc}
        \toprule
        Schedule & FID~\(\downarrow\) & \(\Delta\)FID~\(\uparrow\) \\
        \midrule
        Baseline                         & \(8.60\) & -- \\
        Model-aware, epoch 400 estimate & \(5.57\) & \(3.03\) \\
        Model-aware, epoch 100 estimate & \(5.63\) & \(2.96\) \\
        \bottomrule
    \end{tabular}
\end{table}

The slightly lower DDPM FID obtained with the epoch-100 estimate is not
interpreted as evidence that earlier checkpoints are systematically
preferable; rather, these results indicate that the estimated risk profile
is sufficiently stable across substantially different training stages for
the resulting schedule to remain effective.

\paragraph{Risk-estimation cost.}
We use \(12{,}800\) Monte Carlo samples per reference interval: 199
intervals for DDPM (\(2.55\)M sample-level forward evaluations) and 200
for flow matching (\(2.56\)M). The same respective budgets are used on
CIFAR-10 and ImageNet-64. On ImageNet-64, a full DDPM run at 800k updates
with effective batch size \(128\) processes \(102.4\)M training sample
instances, while a 600k-update flow-matching run processes \(76.8\)M.
Thus risk estimation amounts to only about \(2.5\%\) and \(3.3\%\) of
these sample counts, respectively, and consists only of forward
evaluations. This estimation cost is incurred only once.

\subsection{Flow-Matching Finite-Step Integration Diagnostic}
\label{app:fm-finite-step-diagnostic}

Table~\ref{tab:fm-finite-step-diagnostic} provides the complete finite-step
integration diagnostic underlying the discussion in
Section~\ref{sec:finite-step-integration}. We compare several integrators
over increasing NFE budgets to separate schedule effects from numerical
behavior specific to coarse integration grids.

\begin{table}[t]
    \centering
    \small
    \setlength{\tabcolsep}{12pt}
    \caption{
        Finite-step integration diagnostic for CIFAR-10 flow matching at
        epoch 900. Each entry reports Baseline \(\rightarrow\) Model-aware
        FID~\(\downarrow\) (relative improvement~\(\uparrow\)). Single-seed results. Heun2 (last-mid.)
        replaces the final Heun2 step with a midpoint step to diagnose
        low-NFE terminal-stage sensitivity.
    }
    \label{tab:fm-finite-step-diagnostic}
    \begin{tabular}{lccc}
        \toprule
        Integrator & 16 NFE & 32 NFE & 64 NFE \\
        \midrule
        Midpoint
        & \(7.59 \!\to\! 4.60\;(39.3\%)\)
        & \(5.52 \!\to\! 4.63\;(16.1\%)\)
        & \(4.45 \!\to\! 4.04\;(9.2\%)\) \\
        Heun2
        & \(52.97 \!\to\! 33.66\;(36.5\%)\)
        & \(20.14 \!\to\! 9.95\;(50.6\%)\)
        & \(6.04 \!\to\! 3.76\;(37.8\%)\) \\
        Heun2 (last-mid.)
        & \(7.00 \!\to\! 4.44\;(36.5\%)\)
        & \(5.20 \!\to\! 4.07\;(21.8\%)\)
        & \(4.27 \!\to\! 3.93\;(8.0\%)\) \\
        RK4
        & \(30.95 \!\to\! 12.84\;(58.5\%)\)
        & \(9.55 \!\to\! 4.43\;(53.6\%)\)
        & \(\mathbf{3.73} \!\to\! \mathbf{3.52}\;(5.5\%)\) \\
        Euler
        & \(9.53 \!\to\! 9.98\;(-4.7\%)\)
        & \(6.98 \!\to\! 7.90\;(-13.2\%)\)
        & \(5.47 \!\to\! 5.80\;(-6.0\%)\) \\
        \bottomrule
    \end{tabular}
\end{table}

The unusually poor low-NFE behavior of standard Heun2 is closely tied to
its terminal-stage evaluation. Replacing only the final Heun2 step with a
midpoint step reduces the baseline FID from \(52.97\) to \(7.00\) at
16 NFE and from \(20.14\) to \(5.20\) at 32 NFE. The discrepancy also
shrinks substantially under grid refinement: the baseline gap between
standard and last-midpoint Heun2 decreases from \(45.97\) FID at 16 NFE
to \(1.77\) FID at 64 NFE.

RK4 exhibits a related dependence on grid resolution, with its baseline
FID improving from \(30.95\) at 16 NFE to \(3.73\) at 64 NFE. The latter
is lower than the corresponding midpoint baseline of \(4.45\), and the
model-aware schedule further reduces it to \(3.52\). Euler instead yields
higher FID under the model-aware schedule at all three CIFAR-10 budgets.
Since Euler improves at all three tested budgets on ImageNet-64
(Table~\ref{tab:fm-imagenet-main}), this degradation is setting-specific rather
than a systematic incompatibility with first-order integration.

\subsection{ImageNet-64 Flow-Matching Qualitative Samples}
\label{app:imagenet-fm-qualitative}

\begin{figure*}[t]
    \centering
    \includegraphics[width=0.485\textwidth]{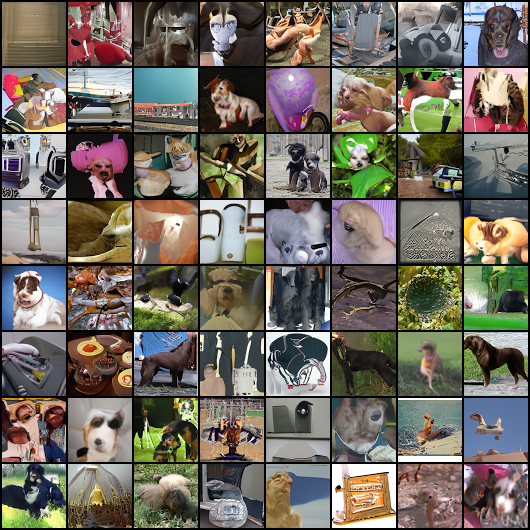}
    \hfill
    \includegraphics[width=0.485\textwidth]{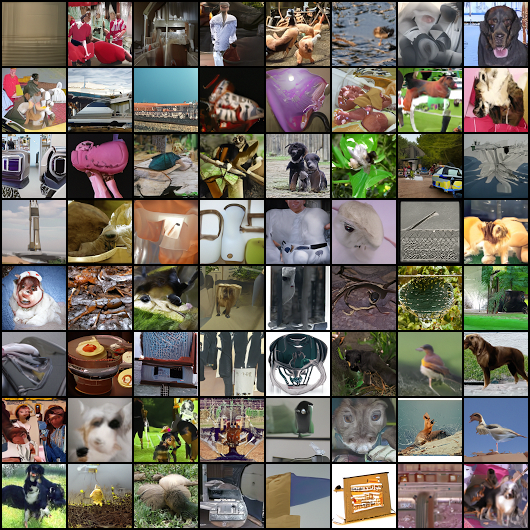}
    \caption{
        Uncurated samples from ImageNet-64 OT-CFM models at the 600k
        checkpoint and \(16\) NFE using midpoint integration.
        \textbf{Left:} standard Cond-OT schedule.
        \textbf{Right:} model-aware reparameterization with \(\lambda=450\).
        Both grids use the same \(64\) fixed sampling seeds.
    }
    \label{fig:imagenet-fm-qualitative}
\end{figure*}

Figure~\ref{fig:imagenet-fm-qualitative} complements the 16-NFE midpoint
result in Table~\ref{tab:fm-imagenet-main}. With the same \(64\) fixed
sampling seeds, the model-aware grid exhibits more coherent object and
scene structure, consistent with the FID reduction from \(41.52\) to
\(30.11\).

\subsection{Risk-Profile and Allocation-Density Agreement}
\label{app:shared-risk-allocation-agreement}

This appendix quantifies and extends the risk-profile and
allocation-density agreement reported in
Section~\ref{sec:shared-risk-allocation}.

\begin{figure}[t]
    \centering
    \includegraphics[
        width=0.68\linewidth
    ]{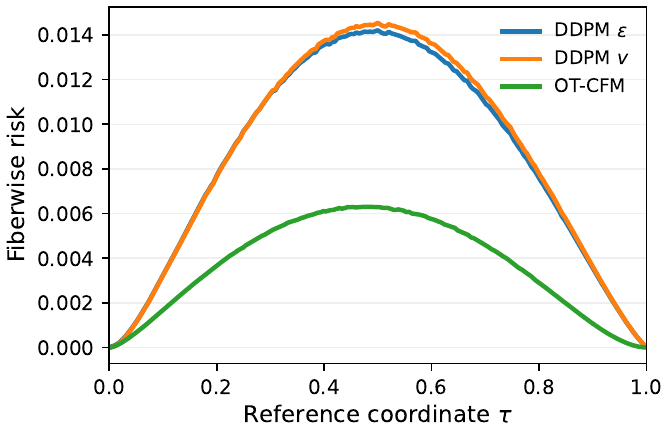}
    \caption{
        Raw fiberwise-risk profiles for representative CIFAR-10
        settings in their respective reference coordinates. Absolute risk
        magnitudes differ substantially across systems.
    }
    \label{fig:fiberwise-risk-profiles}
\end{figure}

\paragraph{Comparison protocol and metrics.}
Figure~\ref{fig:fiberwise-risk-profiles} shows representative CIFAR-10
fiberwise-risk profiles in their respective kinetic reference
coordinates. Their absolute magnitudes differ substantially across systems,
motivating a shape comparison after normalization.

For cross-system shape comparison, we normalize each risk profile to unit
area,
\begin{equation}
\label{eq:unit-area-risk-normalization}
    \bar R(\tau)
    =
    \frac{R(\tau)}
         {\int_0^1 R(u)\,du}.
\end{equation}
Model-aware allocation densities are constructed from the raw risk
profiles with the corresponding \(\lambda\) values, not from \(\bar R\).

To characterize the agreement reported in
Section~\ref{sec:shared-risk-allocation}, we fit the reference
profiles using the epoch-320 DDPM \(\epsilon\)-prediction and
epoch-400 OT-CFM baseline risk profiles on CIFAR-10, together
with their corresponding model-aware allocation densities.
The exponent \(5/4\) and coefficient \(4\) are fitted over simple
candidate values.
Neither parameter is theoretically distinguished; the fitted
profiles are empirical approximations rather than uniquely
identified parametric laws.
The risk-shape reference and analytic template remain fixed
in all subsequent comparisons.

For both normalized-risk and allocation-density comparisons, let \(f\)
denote the empirical profile normalized to unit area and \(g\) the corresponding frozen
reference profile. We
report Pearson correlation together with
\begin{equation}
\label{eq:risk-allocation-shape-metrics}
    E_{L^2}^{\rm rel}(f,g)
    =
    \frac{\|f-g\|_{L^2([0,1])}}
         {\|f\|_{L^2([0,1])}},
    \qquad
    \mathrm{TV}(f,g)
    =
    \frac12
    \|f-g\|_{L^1([0,1])}.
\end{equation}
For risk profiles, \(f=\bar R\) and \(g\) is the normalized
\(\sin^{5/4}(\pi\tau)\) reference; for allocation densities, \(f=w\) and
\(g=w_{\rm ana}\).

To quantify mirror asymmetry independently of either frozen reference profile, we
additionally report
\begin{equation}
\label{eq:mirror-asymmetry}
    A_{\rm TV}(f)
    =
    \frac12
    \int_0^1
    \left|
        f(\tau)-f(1-\tau)
    \right|
    \,d\tau .
\end{equation}
This quantity vanishes for a profile symmetric about \(\tau=1/2\).

\paragraph{Agreement across models and settings.}
Table~\ref{tab:shared-risk-allocation-agreement} compares the normalized
risk profiles and allocation densities with the frozen reference profiles.
\begin{table}[t]
    \centering
    \scriptsize
    \setlength{\tabcolsep}{3.0pt}
    \caption{
        Agreement with the frozen risk-shape reference and frozen analytic
        allocation template, together with mirror asymmetry \(A_{\rm TV}\).
        CIFAR-10 DDPM uses the U-Net backbone unless U-ViT-S/2 is specified;
        unqualified OT-CFM uses uniform training-time sampling.
        Allocation densities use \(\lambda=220\) for diffusion and
        \(\lambda=450\) for flow matching unless otherwise specified. The
        DiT \(\lambda=34.7\) row uses the same risk profile as the preceding row.
        \(E_{L^2}^{\rm rel}\), TV, and \(A_{\rm TV}\) are reported in percent.
    }
    \label{tab:shared-risk-allocation-agreement}
    \begin{tabular}{lrrrrrrrr}
        \toprule
        &
        \multicolumn{4}{c}{Normalized risk}
        &
        \multicolumn{4}{c}{Allocation density}
        \\
        \cmidrule(lr){2-5}
        \cmidrule(lr){6-9}
        Setting
        & Pearson~\(\uparrow\)
        & \(E_{L^2}^{\rm rel}\)~\(\downarrow\)
        & TV~\(\downarrow\)
        & \(A_{\rm TV}\)
        & Pearson~\(\uparrow\)
        & \(E_{L^2}^{\rm rel}\)~\(\downarrow\)
        & TV~\(\downarrow\)
        & \(A_{\rm TV}\)
        \\
        \midrule

        CIFAR-10 DDPM \(\epsilon\), ep.~320
        & \(0.9991\) & \(2.20\) & \(1.12\) & \(1.43\)
        & \(0.9988\) & \(1.98\) & \(0.74\) & \(0.72\) \\

        CIFAR-10 DDPM \(\epsilon\), ep.~100
        & \(0.9993\) & \(3.18\) & \(1.51\) & \(1.40\)
        & \(0.9969\) & \(2.16\) & \(0.66\) & \(0.99\) \\

        CIFAR-10 DDPM \(v\)
        & \(0.9994\) & \(1.72\) & \(0.85\) & \(0.96\)
        & \(0.9985\) & \(1.67\) & \(0.62\) & \(0.72\) \\

        CIFAR-10 DDPM \(\epsilon\), U-ViT-S/2
        & \(0.9995\) & \(1.61\) & \(0.82\) & \(1.22\)
        & \(0.9988\) & \(1.61\) & \(0.58\) & \(0.69\) \\

        \midrule

        CIFAR-10 OT-CFM, ep.~400
        & \(0.9917\) & \(6.39\) & \(3.22\) & \(6.41\)
        & \(0.9909\) & \(3.78\) & \(1.50\) & \(2.91\) \\

        CIFAR-10 OT-CFM, ep.~100
        & \(0.9905\) & \(6.98\) & \(3.54\) & \(7.01\)
        & \(0.9897\) & \(5.48\) & \(2.24\) & \(3.37\) \\

        CIFAR-10 OT-CFM, logit-normal
        & \(0.9906\) & \(6.69\) & \(3.40\) & \(6.70\)
        & \(0.9901\) & \(3.73\) & \(1.51\) & \(3.00\) \\

        CIFAR-10 OT-CFM, U-shaped (RF++)
        & \(0.9921\) & \(6.34\) & \(3.19\) & \(6.33\)
        & \(0.9912\) & \(3.90\) & \(1.55\) & \(2.90\) \\

        CIFAR-10 Independent-CFM
        & \(0.9989\) & \(2.25\) & \(1.08\) & \(1.42\)
        & \(0.9953\) & \(7.12\) & \(2.67\) & \(0.82\) \\

        \midrule

        ImageNet-64 DDPM \(\epsilon\)
        & \(0.9948\) & \(4.92\) & \(2.54\) & \(5.02\)
        & \(0.9944\) & \(2.74\) & \(1.19\) & \(2.17\) \\

        ImageNet-64 OT-CFM
        & \(0.9878\) & \(7.78\) & \(3.96\) & \(7.86\)
        & \(0.9888\) & \(6.94\) & \(2.84\) & \(3.70\) \\

        \midrule

        \multicolumn{9}{l}{\emph{Pretrained conditional latent-model diagnostics}} \\

        DiT-XL/2, ImageNet-256, \(\lambda=220\)
        & \(0.9930\) & \(8.70\) & \(4.37\) & \(5.19\)
        & \(0.5223\) & \(88.00\) & \(36.18\) & \(26.43\) \\

        DiT-XL/2, \(\lambda=34.7\)
        & -- & -- & -- & --
        & \(0.9861\) & \(4.40\) & \(1.44\) & \(2.04\) \\

        InstaFlow 2-RF, COCO 2017, \(\lambda=450\)
        & \(0.9835\) & \(9.92\) & \(4.52\) & \(6.34\)
        & \(0.9884\) & \(16.20\) & \(6.59\) & \(1.14\) \\

        \bottomrule
    \end{tabular}
\end{table}

In our CIFAR-10 DDPM experiments, both profiles retain close agreement
with their respective references across prediction targets,
risk-estimation checkpoints, and the U-Net and U-ViT architectures.
For flow matching, both findings persist across endpoint couplings,
training-time sampling distributions, and risk-estimation checkpoints.
The ImageNet-64 experiments extend this agreement to a second dataset
for both model families.

\paragraph{Pretrained-checkpoint diagnostics.}
The DiT-XL/2 and InstaFlow 2-RF checkpoints evaluated in
Appendix~\ref{app:pretrained-latent-diagnostics} also retain close
normalized-risk agreement with the risk-shape reference.
These are diagnostics without retraining or FID evaluation under a
modified schedule. Under the raw-risk allocation construction,
the previously fixed flow-matching value \(\lambda=450\) also yields close
2-RF allocation agreement, whereas the fixed diffusion value \(\lambda=220\)
does not transfer directly to DiT.

Fitting the DiT raw risk by an affine
rescaling of the \(\sin^{5/4}(\pi\tau)\) shape and applying the
risk-recalibration property of
Appendix~\ref{app:risk-recalibration-properties} predicts \(\lambda=34.7\). At
this value, the DiT allocation attains Pearson \(0.9861\), relative \(L^2\)
error \(4.40\%\), TV \(1.44\%\), and \(A_{\rm TV}=2.04\%\). The predicted
value is close to the one-dimensional TV- and relative-\(L^2\)-optimal values,
\(35.3\) and \(37.6\), respectively. Thus the DiT discrepancy is primarily
an absolute risk-scale shift rather than a failure of the normalized shape.

\paragraph{DDPM coordinate controls.}
We next test how the observed risk-profile agreement depends on the
comparison coordinate. The linear-\(\beta\) DDPM control compares the
same pointwise risk values in native model time and in the practical
VP reference coordinate.
At each native model time \(t\), risk \(R(t)\) is evaluated at the
corresponding coefficient pair \((m_t,s_t)\).
We match this risk value to the cosine-baseline time
with the same coefficient pair.
Unit-area normalization is then performed separately in native model
time and in the practical VP reference coordinate.

A separate check asks whether the practical approximation to the VP
kinetic reference coordinate affects the shape comparisons.
We therefore also recompute the cosine- and linear-DDPM profiles in
the exact VP reference coordinate defined in
Appendix~\ref{app:vp-kinetic-reference-coordinate}.
Table~\ref{tab:ddpm-coordinate-controls} reports both coordinate controls.

\begin{table}[t]
    \centering
    \small
    \setlength{\tabcolsep}{4.0pt}
    \caption{
        CIFAR-10 DDPM risk-shape coordinate controls.
        ``Practical VP reference'' and ``exact VP reference'' denote,
        respectively, the practical approximation to and exact form of the
        VP kinetic reference coordinate.
        \(E_{L^2}^{\rm rel}\) and TV are reported in percent.
    }
    \label{tab:ddpm-coordinate-controls}
    \begin{tabular}{llccc}
        \toprule
        Comparison
        & Coordinate
        & Pearson~\(\uparrow\)
        & \(E_{L^2}^{\rm rel}\)~\(\downarrow\)
        & TV~\(\downarrow\) \\
        \midrule

        Cosine DDPM vs.\ risk-shape reference
        & Practical VP reference
        & \(0.9991\)
        & \(2.20\)
        & \(1.12\) \\

        Cosine DDPM vs.\ risk-shape reference
        & Exact VP reference
        & \(0.9988\)
        & \(2.37\)
        & \(1.10\) \\

        \midrule

        Linear DDPM vs.\ risk-shape reference
        & Native model time
        & \(0.3616\)
        & \(64.38\)
        & \(38.00\) \\

        Linear DDPM vs.\ risk-shape reference
        & Practical VP reference
        & \(0.9994\)
        & \(1.67\)
        & \(0.84\) \\

        Linear DDPM vs.\ risk-shape reference
        & Exact VP reference
        & \(0.9995\)
        & \(1.64\)
        & \(0.75\) \\

        \midrule

        Linear vs.\ cosine DDPM
        & Practical VP reference
        & \(0.9998\)
        & \(0.93\)
        & \(0.45\) \\

        \bottomrule
    \end{tabular}
\end{table}

For the direct linear--cosine comparison, \(E_{L^2}^{\rm rel}\)
uses the \(L^2\) norm of the cosine-DDPM profile as its denominator.

The linear-DDPM profile agrees weakly with the risk-shape reference in
native model time but closely matches it in the practical VP reference
coordinate. The direct linear--cosine comparison is similarly tight,
confirming that the two normalized risk profiles closely agree with
each other rather than only agreeing separately with the risk-shape
reference. Thus the linear-DDPM control recovers the shared risk shape
in the kinetic reference coordinate, whereas its native-time profile
does not exhibit this agreement.

Using the exact rather than the practical VP reference coordinate changes
the cosine- and linear-DDPM comparisons with the risk-shape reference
only marginally. Together with the coordinate-level audit in
Appendix~\ref{app:vp-kinetic-reference-coordinate}, this result supports
using the practical VP reference as a close approximation to the exact
VP kinetic reference coordinate in the experiments.

\paragraph{Post-training allocation stability.}
We also test whether the shared normalized risk shape persists after
model-aware training and whether re-estimating risk changes the
one-shot allocation.
Let \(R_0\) denote the baseline-checkpoint risk used to construct the
one-shot allocation \(w_1\). After training with \(w_1\), we estimate
\(R_1\) from the resulting model-aware checkpoint and compute the
diagnostic allocation \(w_2\) from the same closed-form rule with the
original \(\lambda\). We evaluate \(R_1\) in the same kinetic reference coordinate as \(R_0\),
using the corresponding general-target expression from
Section~\ref{sec:general-linear-target-invariance}. Let \(\Phi_i(\tau)=\int_0^\tau w_i(u)\,du\), \(i\in\{1,2\}\). For OT-CFM, the model-aware predictor \(v_{\theta,1}(x,t)\)
estimates \(dx_t/dt\). At \(t=\Phi_1(\tau)\), its velocity in
the reference coordinate is
\(w_1(\tau)v_{\theta,1}(x_\tau,\Phi_1(\tau))\), giving
\begin{equation}
\label{eq:post-training-fm-reference-risk}
    R_1(\tau)
    =
    \tau^2(1-\tau)^2
    \E\left\|
        w_1(\tau)v_{\theta,1}(x_\tau,\Phi_1(\tau))
        -(\epsilon-x_0)
    \right\|^2,
\end{equation}
where \(x_\tau=(1-\tau)x_0+\tau\epsilon\).

For DDPM, \(R_0\) and \(R_1\) are estimated from the epoch-320 baseline
and epoch-400 model-aware checkpoints, respectively; for OT-CFM, both
checkpoints are at epoch 400. Table~\ref{tab:post-training-allocation-stability}
reports the absolute risk-scale ratio
\((\int_0^1R_1(\tau)\,d\tau)/(\int_0^1R_0(\tau)\,d\tau)\),
together with \(\operatorname{TV}(\bar R_1,\bar R_0)\),
\(\operatorname{TV}(w_2,w_1)\), and
\(\|\Phi_2-\Phi_1\|_\infty\).

\begin{table}[h]
    \centering
    \small
    \setlength{\tabcolsep}{6pt}
    \caption{
        Post-training allocation stability on the two primary CIFAR-10
        settings. TV values are reported in percent.
    }
    \label{tab:post-training-allocation-stability}
    \begin{tabular}{lcccc}
        \toprule
        & \multicolumn{2}{c}{Risk}
        & \multicolumn{2}{c}{Allocation} \\
        \cmidrule(lr){2-3}
        \cmidrule(lr){4-5}
        Setting
        & Scale ratio
        & TV~\(\downarrow\)
        & TV~\(\downarrow\)
        & \(\|\Phi_2-\Phi_1\|_\infty\)~\(\downarrow\) \\
        \midrule
        DDPM \(\epsilon\)-pred.
        & \(0.9988\)
        & \(0.161\)
        & \(0.051\)
        & \(4.57\times10^{-4}\) \\

        OT-CFM
        & \(1.0192\)
        & \(0.532\)
        & \(0.284\)
        & \(1.45\times10^{-3}\) \\
        \bottomrule
    \end{tabular}
\end{table}

The model-aware checkpoints retain normalized risk profiles close to
those of the baseline checkpoints, extending the shared-shape evidence
beyond baseline models. Recomputing the allocations from these
post-training risks changes them only marginally, preserving the shared
allocation deformation and supporting the one-shot construction.
This is a stability diagnostic; the proposed construction itself
remains one-shot.

\paragraph{Interpreting the normalized risk shape.}
Across models and settings, the shared normalized risk shape reflects the combination of
\(s_\tau^2\) and the corresponding noise-prediction MSE, rather than
either alone; for velocity prediction, this MSE uses the induced noise
prediction. In the exact VP reference coordinate,
\(s_\tau^2=\sin^2(\pi\tau/2)\), whereas \(s_\tau^2=\tau^2\) for the
standard Cond-OT parameterization. Both factors are monotone and largest
at the noise endpoint, unlike the risk-shape reference.
For \(0<\tau<1\), dividing that reference by the respective factors
yields different noise-prediction MSE shapes.

\paragraph{Mirror asymmetry.}
In our CIFAR-10 and ImageNet-64 experiments, the tested OT-CFM settings
exhibit systematically stronger mirror asymmetry than DDPM and
Independent-CFM: \(A_{\rm TV}\) ranges from \(6.33\%\) to \(7.86\%\) for
normalized risk and from \(2.90\%\) to \(3.70\%\) for allocation density.

For any symmetric reference \(g\),
\begin{equation}
\label{eq:symmetric-reference-discrepancy-bound}
    \mathrm{TV}(f,g)
    \ge
    \frac12 A_{\rm TV}(f),
\end{equation}
so profile asymmetry places a direct lower bound on its discrepancy from
either symmetric frozen reference profile.
For normalized risk,
the observed OT-CFM TV values lie close to the lower bound
\(A_{\rm TV}/2\); for allocation density, the same lower bound remains
substantial but leaves a larger residual in some settings.

The skew also has a consistent direction across the tested OT-CFM settings.
Their normalized-risk centers of mass,
\begin{equation}
\label{eq:profile-center-of-mass}
    \operatorname{COM}(f)
    =
    \int_0^1 \tau f(\tau)\,d\tau ,
\end{equation}
lie on the data side (\(0.480\)--\(0.483\)), whereas their allocation
centers of mass lie on the noise side (\(0.510\)--\(0.512\)).
This reversal is consistent with the reciprocal-square-root
risk-to-allocation relation, which reduces allocation where the estimated
risk is larger. The model-aware allocations thus retain setting-specific asymmetry
absent from the symmetric analytic template.
The CIFAR-10 symmetrization control suggests that retaining this
asymmetry can improve FID
(Appendix~\ref{app:frozen-analytic-allocation-template-validation}).

\subsection{Frozen Analytic Allocation Template: Functional Validation}
\label{app:frozen-analytic-allocation-template-validation}

We next evaluate whether the frozen analytic allocation template defined in
Section~\ref{sec:shared-risk-allocation} is functionally useful
when used directly as a schedule. Its normalized density is
\begin{equation}
\label{eq:analytic-allocation-template}
    w_{\rm ana}(\tau)
    \propto
    \left(
        1+4\sin^{5/4}(\pi\tau)
    \right)^{-1/2},
    \qquad
    \int_0^1 w_{\rm ana}(\tau)\,d\tau=1.
\end{equation}
Its cumulative map
\begin{equation}
\label{eq:analytic-cumulative-time-map}
    \Phi_{\rm ana}(\tau)
    =
    \int_0^\tau w_{\rm ana}(u)\,du
\end{equation}
defines a fixed monotone reparameterization.

We realize the same analytic template in the native numerical
representation of each model family. Across both CIFAR-10 and ImageNet-64, the primary DDPM realization
integrates \(w_{\rm ana}\) directly at the final 1000-step diffusion
resolution and applies the baseline-preserving cosine-phase construction
of Appendix~\ref{app:model-aware-pipeline}. For both datasets, the primary
flow-matching realization uses the same 200-bin piecewise-linear
reparameterization as the corresponding model-aware schedule. These primary realizations are used for all functional-validation
results reported in
Table~\ref{tab:frozen-analytic-allocation-template-validation}.
On CIFAR-10, we additionally evaluate a 199-bin DDPM realization matched
to the estimated model-aware schedule and a smooth flow-matching realization
obtained from the continuous cumulative map.

To quantify how much of the model-aware improvement is retained, we
define the recovered-gain fraction as
\begin{equation}
\label{eq:recovered-gain-fraction}
    G_{\rm rec}
    =
    \frac{
        \mathrm{FID}_{\rm baseline}
        -
        \mathrm{FID}_{\rm analytic}
    }{
        \mathrm{FID}_{\rm baseline}
        -
        \mathrm{FID}_{\rm model\text{-}aware}
    }.
\end{equation}
We leave this ratio unclipped, so \(G_{\rm rec}>1\) indicates that the
analytic schedule attains a lower FID than the corresponding model-aware
schedule.

Table~\ref{tab:frozen-analytic-allocation-template-validation} reports the
corresponding CIFAR-10 and ImageNet-64 results. The analytic template
recovers \(98.1\%\) and \(98.0\%\) of the model-aware
improvement for CIFAR-10 DDPM \(\epsilon\)-prediction and \(v\)-prediction,
respectively, and \(75.8\%\) in the matched 400-epoch OT-CFM setting. On
ImageNet-64, it attains \(109.7\%\) gain recovery for DDPM and \(76.2\%\) for
600k OT-CFM.

These results clarify that the analytic template does not supplant the
model-aware formulation. For DDPM, its near-complete gain recovery shows
that the shared allocation deformation accounts for most of the
model-aware FID improvement. However, the analytic template is not an independently motivated heuristic
schedule. The cross-system risk--allocation analysis identifies the shared
normalized risk shape, while the fixed-curve formulation in
Section~\ref{sec:fixed-curve-model-aware-allocation} yields the
analytic template's reciprocal-square-root dependence on risk. For flow matching, the
\(75.8\%\) and \(76.2\%\) recovery fractions show that the full model-aware
construction retains substantial additional gains from setting-specific
risk structure.

\begin{table}[t]
    \centering
    \small
    \setlength{\tabcolsep}{12pt}
    \caption{
        FID~\(\downarrow\) and model-aware gain recovery of the frozen analytic
        allocation template on CIFAR-10 and ImageNet-64. The CIFAR-10 OT-CFM row uses the matched 400-epoch
        setting; the ImageNet-64 DDPM and OT-CFM rows use the 800k and 600k
        checkpoints, respectively. Single-seed results. Gain recovery is
        measured relative to the improvement from the baseline to the
        model-aware schedule and computed from unrounded FIDs.
    }
    \label{tab:frozen-analytic-allocation-template-validation}
    \begin{tabular}{lcccc}
        \toprule
        Setting
        & FID (Baseline)~\(\downarrow\)
        & FID (Model-aware)~\(\downarrow\)
        & FID (Analytic)~\(\downarrow\)
        & \(G_{\rm rec}\)~\(\uparrow\) \\
        \midrule
        CIFAR-10 DDPM \(\epsilon\)-pred.
        & \(9.43\)
        & \(7.90\)
        & \(7.93\)
        & \(98.1\%\) \\
        CIFAR-10 DDPM \(v\)-pred.
        & \(11.31\)
        & \(9.65\)
        & \(9.68\)
        & \(98.0\%\) \\
        CIFAR-10 OT-CFM
        & \(8.60\)
        & \(5.57\)
        & \(6.30\)
        & \(75.8\%\) \\
        \midrule
        ImageNet-64 DDPM \(\epsilon\)-pred.
        & \(25.01\)
        & \(23.26\)
        & \(23.09\)
        & \(109.7\%\) \\
        ImageNet-64 OT-CFM, 600k
        & \(41.52\)
        & \(30.11\)
        & \(32.83\)
        & \(76.2\%\) \\
        \bottomrule
    \end{tabular}
\end{table}

For CIFAR-10 flow matching, we further examine whether the remaining gap
to the model-aware schedule is explained primarily by the setting-specific
asymmetry visible in both panels of
Figure~\ref{fig:shared-risk-allocation}: in the normalized-risk profile on
the left and the induced allocation density on the right. We symmetrize the \(\lambda=450\) model-aware schedule by pairwise
averaging its 200 interval allocations:
\(\Delta t_k^{\rm sym}
=(\Delta t_k+\Delta t_{199-k})/2\), \(k=0,\ldots,199\).
The resulting OT-CFM schedule attains an FID of \(5.84\),
compared with \(5.57\) for the original model-aware schedule and \(6.30\)
for the analytic template. Relative to the symmetrized
model-aware schedule, the analytic template recovers \(83.1\%\)
of the FID improvement over the baseline. Symmetrizing the model-aware
allocation therefore narrows the gap, but does not account for it
completely.

Finally, using the CIFAR-10 settings, we test whether the observed benefit
is sensitive to the numerical realization of the same analytic template.
For DDPM \(\epsilon\)-prediction, the direct 1000-step realization gives an
FID of \(7.93\), compared with \(8.05\) for the 199-bin realization. For
OT-CFM, the 200-bin piecewise-linear realization gives \(6.30\), compared
with \(6.47\) for the smooth realization. All four remain substantially
better than their respective baselines, so we view these differences as
numerical realization effects rather than evidence for a generally
preferred discretization.

\subsection{Empirical Bayes Decomposition of Fiberwise Risk}
\label{app:empirical-bayes-fiberwise-risk}

Independent-CFM is included in the coupling-robustness experiments of
Section~\ref{sec:fm-coupling-time-sampling}, and its normalized risk and
allocation shapes are included in the unified comparisons of
Appendix~\ref{app:shared-risk-allocation-agreement}. Here we use the
independent endpoint structure of DDPM and Independent-CFM to ask which
part of the fiberwise-risk profile carries the shared broad shape.

Using this independent endpoint structure, we apply an empirical Bayes
decomposition along the corresponding reference coordinate \(\tau\). Let
\begin{equation}
\label{eq:empirical-data-distribution}
    p_{\rm emp}(x_0)
    =
    \frac{1}{50{,}000}
    \sum_{j=1}^{50{,}000}
    \delta(x_0-x_j)
\end{equation}
be the finite empirical CIFAR-10 training distribution. In this subsection,
outer expectations are over \(x_0\sim p_{\rm emp}\) and independent
\(\epsilon\sim\mathcal N(0,I)\), whereas
\(\E_{\rm emp}[\cdot\mid x_\tau]\) denotes the corresponding
empirical-posterior conditional expectation. For the affine
observation \(x_\tau=m_\tau x_0+s_\tau\epsilon\), the posterior mean of \(x_0\) is
\begin{equation}
\label{eq:empirical-posterior-mean}
    \E_{\rm emp}[x_0\mid x_\tau=x]
    =
    \frac{
        \sum_j x_j
        \exp\!\left(
            -\|x-m_\tau x_j\|^2/(2s_\tau^2)
        \right)
    }{
        \sum_j
        \exp\!\left(
            -\|x-m_\tau x_j\|^2/(2s_\tau^2)
        \right)
    }.
\end{equation}

For the fiberwise risk \(R\), the standard squared-error Bayes
decomposition gives
\begin{equation}
\label{eq:empirical-bayes-risk-decomposition}
    R(\tau)=P(\tau)+M(\tau),
\end{equation}
where \(P\) is the Bayes-risk component under the empirical data
distribution \(p_{\rm emp}\) and \(M\) is the Bayes-excess component.
For a fixed empirical data distribution and
probability path, \(P\) is independent of the trained predictor, whereas
\(M\) measures the predictor-dependent excess risk above the empirical
Bayes predictor.

For DDPM \(\epsilon\)-prediction,
\begin{equation}
\label{eq:ddpm-bayes-risk-components}
    \begin{aligned}
    R_{\rm DDPM}(\tau)
    &=
    (1-\bar\alpha_\tau)
    \E\left\|
        \epsilon-\epsilon_\theta(x_\tau,\tau)
    \right\|^2 ,\\
    P_{\rm DDPM}(\tau)
    &=
    \bar\alpha_\tau
    \E
    \left\|
        x_0-\E_{\rm emp}[x_0\mid x_\tau]
    \right\|^2 .
    \end{aligned}
\end{equation}
For Independent-CFM,
\begin{equation}
\label{eq:independent-cfm-path}
    x_\tau=(1-\tau)x_0+\tau\epsilon,
\end{equation}
and
\begin{equation}
\label{eq:independent-cfm-bayes-risk-components}
\begin{aligned}
    R_{\rm ICFM}(\tau)
    &= \tau^2(1-\tau)^2
       \E\left\|
           v_\theta(x_\tau,\tau)-(\epsilon-x_0)
       \right\|^2,\\
    P_{\rm ICFM}(\tau)
    &= (1-\tau)^2
       \E\left\|
           x_0-\E_{\rm emp}[x_0\mid x_\tau]
       \right\|^2.
\end{aligned}
\end{equation}
In both cases,
\begin{equation}
\label{eq:bayes-excess-risk}
    M(\tau)=R(\tau)-P(\tau).
\end{equation}
The posterior means are evaluated against the same finite empirical
training distribution, and the decomposition uses the same reference-bin risk estimates used for schedule construction.

Table~\ref{tab:empirical-bayes-fiberwise-risk} summarizes the integrated
component shares and the agreement of their normalized shapes with the
total fiberwise-risk profile.

\begin{table}[t]
    \centering
    \small
    \setlength{\tabcolsep}{12pt}
    \caption{
        Bayes-risk/Bayes-excess decomposition of the CIFAR-10
        fiberwise-risk profiles. Integrated shares are computed under the
        finite empirical training distribution. Shape metrics compare
        unit-area profiles.
    }
    \label{tab:empirical-bayes-fiberwise-risk}
    \begin{tabular}{lccccc}
        \toprule
        Setting
        & \(\frac{\int_0^1 P}{\int_0^1 R}\) (\%)
        & \(\frac{\int_0^1 M}{\int_0^1 R}\) (\%)
        & Pearson\((R,P)\)
        & Pearson\((R,M)\)
        & \(\mathrm{TV}(R,M)\) (\%) \\
        \midrule
        DDPM \(\epsilon\)-pred.
        & \(3.68\)
        & \(96.32\)
        & \(-0.4333\)
        & \(0.9923\)
        & \(3.25\) \\
        Independent-CFM
        & \(6.26\)
        & \(93.74\)
        & \(-0.3514\)
        & \(0.9809\)
        & \(5.57\) \\
        \bottomrule
    \end{tabular}
\end{table}

The Bayes-excess component accounts for \(96.32\%\) of the integrated
DDPM risk and \(93.74\%\) of the Independent-CFM risk. Its normalized
shape also closely follows the total risk in both settings, with Pearson
correlations of \(0.9923\) and \(0.9809\) and TV discrepancies of \(3.25\%\)
and \(5.57\%\), respectively. By contrast, the Bayes-risk component
has weakly negative shape correlation with the total risk. Thus the broad
risk-profile shape shared across these two independently coupled systems
is carried predominantly by the Bayes-excess component---the part that
depends on the learned predictor---rather than by the Bayes-risk
component alone.

This decomposition is deliberately interpreted only as a diagnostic of the
empirical fiberwise risk used for schedule construction. Because
\(p_{\rm emp}\) is the finite CIFAR-10 training set, the empirical Bayes
quantities above are not estimates of population-level intrinsic image
uncertainty; in particular, the empirical posterior can become highly
concentrated in low- and moderate-noise regions.

\end{document}